\documentclass[10pt,journal,compsoc]{IEEEtran}

\usepackage{amsmath,amsfonts,amssymb}
\usepackage{algorithmic}
\usepackage{algorithm}
\usepackage{array}
\usepackage[caption=false,font=normalsize,labelfont=sf,textfont=sf]{subfig}
\usepackage{textcomp}
\usepackage{stfloats}
\usepackage{url}
\usepackage{verbatim}
\usepackage{graphicx}
\usepackage{caption}
\usepackage{hyperref}

\def\exp{\text{exp\,}}
\def\eg{\emph{e.g.}\,}
\def\ie{\emph{i.e.}\,}

\usepackage{tikz}
\usepackage{comment}
\usepackage{color}

\usepackage{multirow}
\usepackage{booktabs}
\usepackage{arydshln}
\usepackage{marvosym}
\newcommand{\ra}[1]{\renewcommand{\arraystretch}{#1}}

\ifCLASSOPTIONcompsoc
  \usepackage[nocompress]{cite}
\else
  \usepackage{cite}
\fi
\ifCLASSINFOpdf
\else
\fi
\begin{document}
%
\title{Multi-Granularity Prediction with Learnable Fusion for Scene Text Recognition}
%
%
%
%

\author{Cheng~Da$\,^{\ast}$,
        Peng~Wang$\,^{\ast}$,
        and~Cong~Yao \textsuperscript{\Letter} 
\IEEEcompsocitemizethanks{\IEEEcompsocthanksitem C. Da, P. Wang and C. Yao are with Alibaba DAMO Academy, Beijing, 100102, China.\protect\\
E-mail: \{dc.dacheng08, wdp0072012, yaocong2010\}@gmail.com
\IEEEcompsocthanksitem $\,^{\ast}$ Equal contribution. \Letter~Corresponding author: Cong Yao.
}
}

%
%

\markboth{Journal of \LaTeX\ Class Files}%
{Shell \MakeLowercase{\textit{et al.}}: Bare Advanced Demo of IEEEtran.cls for IEEE Computer Society Journals}
%



\IEEEtitleabstractindextext{%
\begin{abstract}
Due to the enormous technical challenges and wide range of applications, scene text recognition (STR) has been an active research topic in computer vision for years. To tackle this tough problem, numerous innovative methods have been successively proposed, and incorporating linguistic knowledge into STR models has recently become a prominent trend. In this work, we first draw inspiration from the recent progress in Vision Transformer (ViT) to construct a conceptually simple yet functionally powerful vision STR model, which is built upon ViT and a tailored Adaptive Addressing and Aggregation (A$^3$) module. It already outperforms most previous state-of-the-art models for scene text recognition, including both pure vision models and language-augmented methods. To integrate linguistic knowledge, we further propose a Multi-Granularity Prediction strategy to inject information from the language modality into the model in an implicit way, \ie, subword representations (BPE and WordPiece) widely used in NLP are introduced into the output space, in addition to the conventional character level representation, while no independent language model (LM) is adopted. To produce the final recognition results, two strategies for effectively fusing the multi-granularity predictions are devised: the first strategy, called Confidence-based Fusion Strategy (CFS), employs a straightforward rule based on the confidence scores of the multi-granularity predictions, while the second one, called Learnable Fusion Strategy (LFS), is realized with a trainable module to directly measure the cross-modal similarities between text (predicted words) and images, akin to CLIP. The resultant algorithm (termed MGP-STR) is able to push the performance envelope of STR to an even higher level. Specifically, MGP-STR with LFS achieves an average recognition accuracy of $94\%$ on standard benchmarks for scene text recognition (such as IIIT 5K-word, ICDAR 2015, SVT and CUTE). Moreover, it also achieves state-of-the-art results on widely-used handwritten benchmarks (IAM, CVL and RIMES) as well as more challenging scene text datasets (ArT, COCO-Text and Uber-Text), demonstrating the generality of the proposed MGP-STR algorithm. The source code and models will be available at: \url{https://github.com/AlibabaResearch/AdvancedLiterateMachinery/tree/main/OCR/MGP-STR}.
\end{abstract}

\begin{IEEEkeywords}
Scene Text Recognition, Multi-Granularity Prediction, Handwritten Text Recognition, Vision Transformer.
\end{IEEEkeywords}}

\maketitle

\IEEEdisplaynontitleabstractindextext

%
\IEEEpeerreviewmaketitle

\ifCLASSOPTIONcompsoc
\IEEEraisesectionheading{\section{Introduction}\label{sec:introduction}}
\else
\section{Introduction}
\label{sec:introduction}
\fi

%
%
%
%
\IEEEPARstart{R}{eading} text from natural scenes is definitely one of the most indispensable capabilities when building an automated machine with high-level intelligence, as it could provide rich and precise semantic information~\cite{long2021scene}. This explains the reason why researchers from the computer vision community have sedulously explored and investigated this complex and challenging task for decades~\cite{long2021scene,zhu2016scene,chen2021text}. Scene text recognition (STR) involves decoding textual content from natural images (usually cropped sub-images), which is a key component in text reading pipelines.

\begin{figure}[!htp]\centering
 \includegraphics[width=0.45\textwidth]{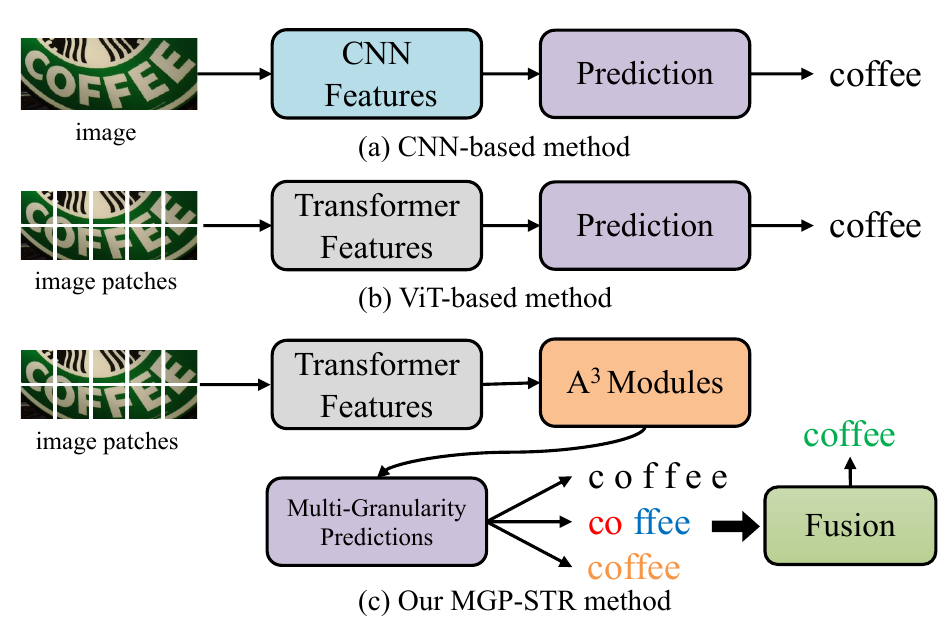}
 \caption{Pipelines of classic CNN-based, ViT-based and the proposed MGP-STR scene text recognition methods are illustrated in (a), (b) and (c), respectively. (Best viewed in color.)}
 \label{fig:pipeline}
\vspace{-4mm}
\end{figure}

Previously, a number of methods~\cite{CRNN,Focusing,ASTER,MASTER} have been proposed to address the problem of scene text recognition mainly from the perspective of \textbf{\textit{vision}}. Recently, there emerges a new trend that \textbf{\textit{linguistic knowledge}} is introduced into the text recognition process, to enhance the performance on difficult cases in real-world scenarios. Concretely, SRN~\cite{SRN} devised a global semantic reasoning module (GSRM) to model global semantic context. ABINet~\cite{ABInet} proposed Bidirectional Cloze Network (BCN) as the language model to learn bidirectional feature representation. MATRN~\cite{matrn} made a further step by introducing multi-modal feature enhancements with bi-directional fusions.
These STR methods, which \textbf{\textit{explicitly}} adopt an independent and separate language model to capture rich language prior, have proven to achieve considerable performance improvements.

\begin{figure}[t]\centering
 \includegraphics[width=0.45\textwidth]{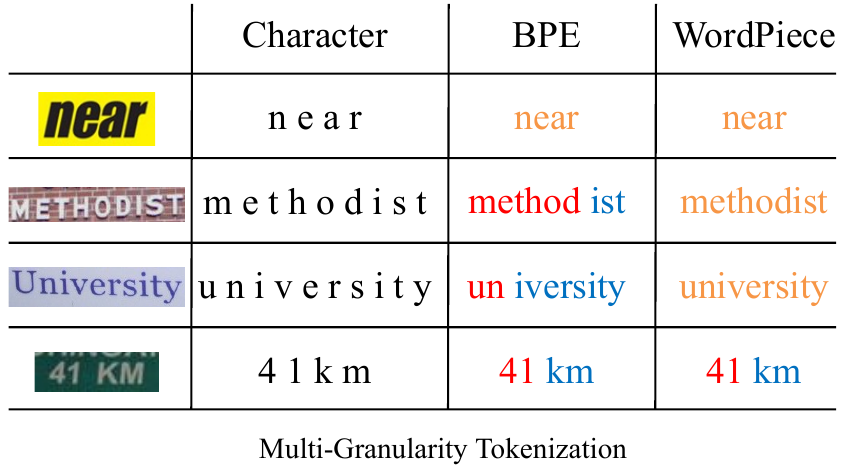}
 \caption{Examples of Character, BPE and WordPiece subword representations. (Best viewed in color.)}
 \label{fig:mgp}
\vspace{-3mm}
\end{figure}

In this paper, we propose to integrate linguistic knowledge in an \textbf{\textit{implicit}} way for scene text recognition. Specifically, we first construct a pure vision STR model based on ViT~\cite{dosovitskiy2020image} and a tailored Adaptive Addressing and Aggregation (A$^3$) module inspired by TokenLearner~\cite{tokenlearner}. This model serves as a strong baseline, which already achieves better performance than previous methods for scene text recognition, according to the experimental comparisons. To further make use of linguistic knowledge to augment this vision STR model, we explore a Multi-Granularity Prediction (MGP) strategy to inject information from the language modality into the vision model. The output space of the model is expanded, in which subword representations (BPE and WordPiece) are introduced, \ie, the augmented model would produce two extra subword-level predictions, besides the original character-level prediction. Fig.~\ref{fig:mgp} depicts several examples where words with multi-granularity representations are illustrated. 

Notably, the proposed method does not include an explicit and separate language model, in contrast to SRN~\cite{SRN} and ABINet~\cite{ABInet}. In the training phase, the resultant model (named MGP-STR) is optimized with a standard multi-task learning paradigm (three losses for three types of predictions), and the linguistic knowledge is naturally integrated into the ViT-based STR model through training. In the inference phase, the three types of predictions will be fused to give the final recognition results. Experiments on standard benchmarks verify that the proposed MGP-STR algorithm can obtain state-of-the-art performance. Another advantage of MGP-STR is that it does not involve iterative refinement, which could be time-consuming in the inference phase~\cite{ABInet,SRN}. The pipeline of the proposed MGP-STR algorithm as well as those of previous CNN-based and ViT-based methods are shown in Fig.~\ref{fig:pipeline}. In a nutshell, the major difference between MGP-STR and other STR methods lies in that MGP-STR generates three types of predictions, representing textual information at different granularities: from individual characters to short character combinations, and even whole words, while in most previous STR methods the predictions are usually at character level (i.e., single-granularity).

The MGP-STR algorithm was originally proposed in our previous conference work~\cite{mgp}, and this paper extends that article with the following improvements and modifications: 
(1) To better unleash the potential of the MGP-STR algorithm, we design a novel fusion strategy (see Sec.~\ref{sec:learnable}), called Learnable Fusion Strategy (LFS), which is realized with a trainable module built on top of the original MGP-STR model, going beyond simple rule-based fusion means. Experiments demonstrate that the proposed LFS can substantially boost the recognition accuracy on standard benchmarks, while only bringing a marginal increase in computational burden.
(2) We conduct additional experiments and comparisons to further validate the effectiveness and applicability of the proposed MGP-STR algorithm (see Sec.~\ref{Sec:sota}). Concretely, we evaluate MGP-STR on more datasets with handwritten text (IAM, CVL and RIMES) and scene text (ArT, COCO-Text and Uber-Text), and find that MGP-STR can also achieve state-of-the-art performances on these challenging benchmarks. 
(3) We verify various recent ViT backbone models (such as those from MAE, DINO and BLIP) and show that, once adequately trained with enough domain-specific data (text images in this work), these ViT models can work equally well on the task of scene text recognition (see Sec.~\ref{sec:ViTBackbones}).
(4) We provide more technical details, in-depth analyses and discussions regarding MGP-STR and point out potential directions for future research (see Sec.~\ref{Sec:sota} and Sec.~\ref{sec:discussion}).


In summary, the main contributions of this work are as follows:
\begin{itemize}
  \item We first construct a pure vision STR model, which is built upon a Vision Transformer (ViT) backbone and a specially designed Adaptive Addressing and Aggregation (A$^3$) module. This model, as a strong baseline, already outperforms previous methods for scene text recognition.

  \item To explore an alternative way for incorporating linguistic knowledge in STR, we introduce subword representations (BPE and WordPiece) to facilitate multi-granularity prediction, and prove that implicitly utilizing linguistic knowledge can be very effective as well and therefore an independent language model (as used in SRN and ABINet) is not indispensable for STR models.

  \item To produce the final recognition results, we devise two effective strategies to fuse the three types of outputs (Character, BPE and WordPiece). The first strategy, called Confidence-based Fusion Strategy (CFS), employs a straightforward rule relying on the confidences of the outputs, while the second one, called Learnable Fusion Strategy (LFS), is trained to directly measure the cross-modal similarities between textual sequences and images, inspired by CLIP~\cite{CLIP}.

  \item Extensive experiments demonstrate that the proposed MGP-STR algorithm achieves state-of-the-art performances on various text recognition benchmarks. Specifically, MGP-STR with LFS achieves an average recognition accuracy of $94\%$ on six standard benchmarks (IC13~\cite{IC13}, SVT~\cite{SVT}, IIIT~\cite{IIIT}, IC15~\cite{IC15}, SVTP~\cite{SVTP} and CUTE~\cite{CUTE}). Besides, it also achieves state-of-the-art results on standard handwritten text datasets (IAM~\cite{IAM}, CVL~\cite{CVL} and RIMES~\cite{RIMES}) and more challenging scene text benchmarks (ArT~\cite{ArT}, COCO-Text~\cite{COCO} and Uber-Text~\cite{Uber}).
  
\end{itemize}

\section{Related Work}

We will briefly review existing ideas and methods that are highly related to our work. For more details, please refer to previous survey papers~\cite{long2021scene,zhu2016scene,priya2016online,purohit2016literature,chen2023vlp}.

\subsection{Scene Text Recognition}

Scene Text Recognition (STR) is a long-term subject of attention and research~\cite{zhu2016scene,long2021scene,chen2021text}. With the popularity of deep learning methods~\cite{vgg,resnet,rnn1}, its effectiveness in the field of STR has been extensively verified. Depending on whether linguistic information is applied, we roughly divide STR methods into two categories, \ie, language-free and language-augmented methods.

\subsubsection{Language-Free STR Methods}

The mainstream way for image feature extraction in STR methods is CNN~\cite{vgg,resnet}.
For example, previous STR methods~\cite{CRNN,rare,rnn1} utilize VGG, and current STR methods~\cite{Rosetta,STAR-Net,deep,GCRNN} employ ResNet~\cite{resnet} for better performance.
Based on the powerful CNN features, various methods~\cite{zhang2022context,PIMNet,liu2022perceiving} are proposed to tackle the STR problem.
CTC-based methods~\cite{CRNN,wan20192d,STAR-Net,gtc,ctc2} use the Connectionist Temporal Classification (CTC)~\cite{ctc} to accomplish sequence recognition.
Segmentation-based methods~\cite{seg,Vocabulary,TextSpotter,TextScanner} cast STR as a semantic segmentation problem. 

Inspired by the great success of Transformer~\cite{trans} in natural language processing (NLP) tasks, the application of Transformer in STR has also attracted more attention~\cite{ViTSTR,ptie,TrOCR}. Vision Transformer (ViT)~\cite{dosovitskiy2020image} that directly processes image patches without convolutions opens the beginning of using Transformer blocks instead of CNNs to solve computer vision problems~\cite{liu2021swin,SegFormer}, leading to prominent results. ViTSTR~\cite{ViTSTR} attempts to simply leverage the feature representations of the last layer of  ViT for parallel character decoding. PTIE~\cite{ptie} is a transformer model that can process multiple patch resolutions and decode in both the original and reverse character orders. In general, language-free methods often fail to recognize low-quality images due to the lack of language information.

\subsubsection{Language-Augmented STR Methods}

Obviously, language information is favorable to the recognition of low-quality images.
RNN-based methods~\cite{CRNN,rnn1,GCRNN}  can effectively capture the dependency between sequential characters, which can be regarded as the methods with an implicit language model. 
However, they cannot execute decoding in parallel during training and inference.

Recently, Transformer blocks are introduced into CNN-based framework to facilitate language content learning. 
SRN~\cite{SRN} proposes a Global Semantic Reasoning Module (GSRM) to capture the global semantic context through multiple parallel transmissions.
ABINet~\cite{ABInet} presents a Bidirectional Cloze Network (BCN) to explicitly model the language information, which is further used for iterative correction.
VisionLAN~\cite{vlan} proposes a visual reasoning module that simultaneously captures visual and language information by masking input images at the feature level. 
LevOCR~\cite{levocr} and MATRN~\cite{matrn} propose cross-modal transformer modules to explore the interactions between vision and language features for better recognition performances.
PARSeq~\cite{PARSeq} learns an ensemble of internal language modules with shared architecture and weights using a transformer decoder, and it still retains the ability of iterative refinement like ABINet~\cite{ABInet}.

Generally, most works\cite{ABInet,PARSeq,matrn} mentioned above utilize a specific module to capture semantic information at the character level, while we manage to directly utilize multi-granularity semantic information (characters, subwords and even whole words) for STR in this paper.

\subsection{Handwritten Text Recognition}

Recognizing handwritten text remains a significant challenge due to the diversity of writing styles that exist among individuals~\cite{AFDM,DAN,learnToAug}. Handwritten text is a fundamental mode of communication that can take many forms, including memos, whiteboards, handwritten notes, stylus input, postal automation, and reading aids for the visually impaired. The objective of handwritten text recognition is to convert handwritten text to digital content, making textual information more easily accessible.

Contemporary handwritten text recognition methods leverage the recent progress made in deep learning, with many of them drawing inspiration from the architecture of CNNs and RNNs. For instance, \cite{cnn-n-gram} utilizes a CNN to estimate the n-gram profile of an image and match it with the profile of a pre-existing word from a dictionary. \cite{phocnet} extends the latter by incorporating a pyramidal histogram of characters (PHOC), which is primarily used for word spotting. \cite{offline} employs a sequence-to-sequence architecture to encode the visual features and then decodes the sequence of characters in the handwritten text image. \cite{dutta2018improving} proposes an adapted CNN-RNN hybrid architecture with a main focus on effective training for enhancing handwritten recognition.

In recent years, generative modeling has gained attention in the field of handwritten text recognition due to the emergence of adversarial learning and GANs~\cite{GAN}. Several studies have explored the use of synthetic data to augment limited datasets and improve the realism of the data. AFDM~\cite{AFDM} employs adversarial learning to enhance the data in a high-dimensional convolutional feature space, resulting in a robust and invariant word-retrieval detector that can handle all types of variations present in natural images of handwritten text.~\cite{learnToAug} proposes a novel data augmentation method specifically designed for the augmentation of sequence-like characters for text recognition. Additionally, ScrabbleGAN~\cite{scrabblegan} and SLOGAN~\cite{SLOGAN} synthesize handwritten text images that are versatile in both style and lexicon, thereby increasing the robustness of the recognizer.

In this work, we simply apply our MGP-STR algorithm to handwritten text recognition and find that MGP-STR performs quite well on this difficult task.

\vspace{-2mm}
\subsection{Contrastive Learning for Text Recognition}

Self-supervised representation learning techniques utilizing contrastive learning have shown promising improvements in various computer vision applications~\cite{simple, momentum, unsupervised, big}. More recently, these contrastive learning methods have been applied to the field of text recognition. For example, SeqCLR~\cite{seqclr} introduces a sequence-to-sequence contrastive learning framework specifically designed for text recognition. Additionally, PerSec~\cite{liu2022perceiving} presents hierarchical contrastive learning that drives each element of features at high and low levels. DiG~\cite{DiG} integrates contrastive learning and masked image modeling in one self-supervised method to learn discrimination and generation, and utilizes large-scale unlabeled text images for model pre-training. 

In this paper, drawing inspiration from CLIP~\cite{CLIP}, we adopt contrastive learning for the cross-modal matching of text sequences (predicted words) and images, resulting in a Learnable Fusion Strategy that leads to higher text recognition performance.


\section{Methodology}

\begin{figure*}[t]\centering
 \includegraphics[width=0.80\textwidth]{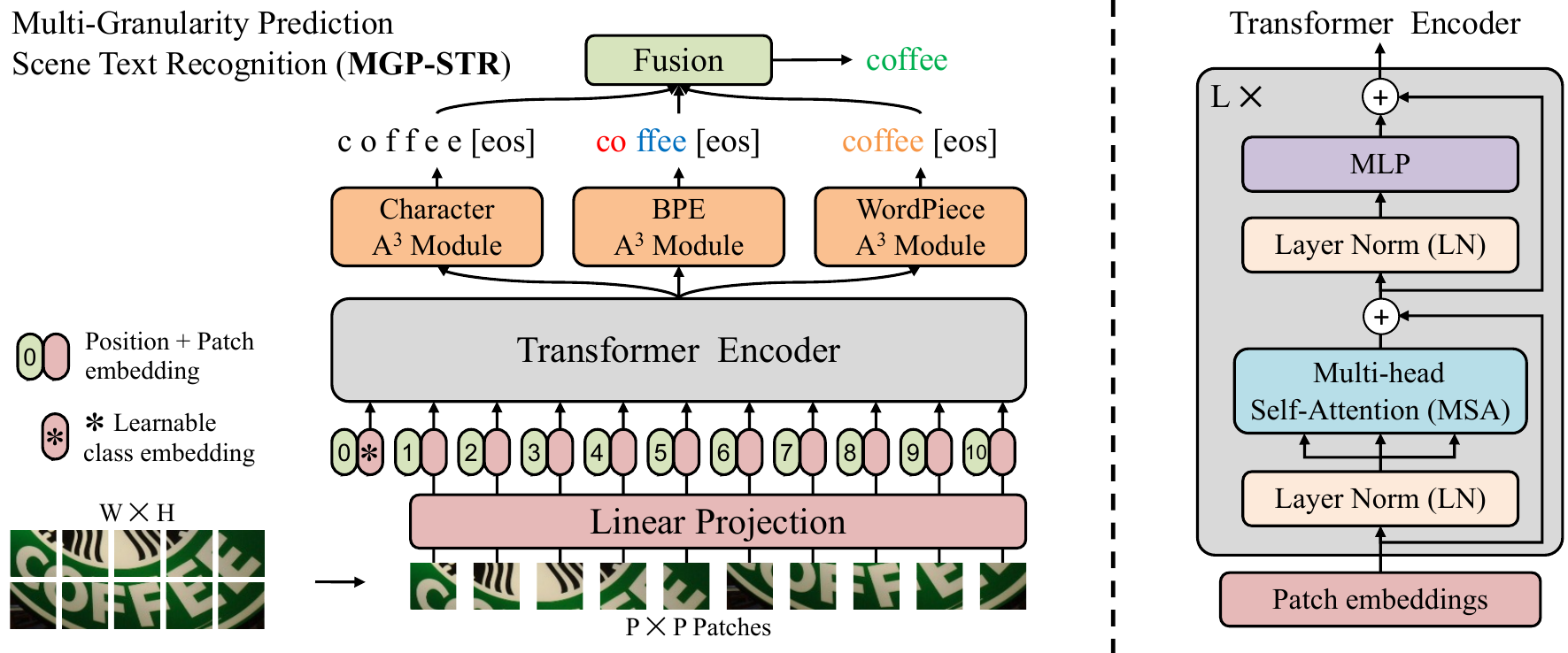}
 \caption{The architecture of the proposed MGP-STR algorithm. }
 \label{fig:overview}
\end{figure*}


The schematic overview of the proposed MGP-STR method is depicted in Fig.~\ref{fig:overview}, which is mainly built upon the original Vision Transformer (ViT) model~\cite{dosovitskiy2020image} (Sec.~\ref{Sec:ViT}). We propose a tailored Adaptive Addressing and Aggregation (A$^3$) module to select a meaningful combination of tokens from ViT and integrate them into one output token corresponding to a specific character, denoted as Character A$^3$ module (Sec.~\ref{sec:a3modules}). Moreover, subword classification heads based on BPE A$^3$ module and WordPiece A$^3$ module are devised for subword predictions, such that the language information can be implicitly modeled (Sec.~\ref{sec:MGP}). Finally, these multi-granularity predictions are merged via a fusion stage, which can be realized with two specially designed fusion strategies: Confidence-based Fusion Strategy (Sec.~\ref{sec:fuse}) and Learnable Fusion Strategy (Sec.~\ref{sec:learnable}).

\subsection{Vision Transformer Backbone}
\label{Sec:ViT}
The fundamental architecture of MGP-STR is Vision Transformer~\cite{dosovitskiy2020image,deit}, where the original image patches are directly utilized for image feature extraction by linear projection. 
As shown in Fig.~\ref{fig:overview}, an input RGB image  $ \mathbf{x} \in  \mathbb{R}^{H \times W \times C} $  is split into non-overlapping patches.
Concretely, the image is reshaped  into a sequence of flattened 2D patches $ \mathbf{x}_p \in  \mathbb{R}^{N \times (P^2 C)} $,
where $(P \times P) $ is the resolution of each image patch and $(P^2 C)$ is the number of feature channels of $ \mathbf{x}_p$. 
In this way, a 2D image is represented as a sequence with $N = HW/P^2$ tokens, which serve as the effective input sequence of Transformer blocks.
Then, these tokens of $ \mathbf{x}_p$  are linearly transcribed into $D$ dimension patch embeddings. 
Similar to the original ViT~\cite{dosovitskiy2020image} backbone, a learnable  $[class]$ token embedding with $D$ dimension is introduced into patch embeddings.
And position embeddings are also added to each patch embedding to retain the positional information,
where the standard learnable 1$D$ position embedding is employed.
Thus, the generation of patch embedding vector is formulated as follows:
\begin{equation}
\begin{split}
\mathbf{z}_0=[\mathbf{x}_{class}; \mathbf{x}^1_p\mathbf{E}; \mathbf{x}^2_p\mathbf{E}; \ldots; \mathbf{x}^N_p\mathbf{E}] + \mathbf{E}_{pos},
\end{split}
\end{equation}
where $\mathbf{x}_{class} \in \mathbb{R}^{ 1 \times D}$ is the $[class]$ embedding,  $\mathbf{E}  \in \mathbb{R}^{ (P^2 C) \times D } $ is a linear projection matrix and  $ \mathbf{E}_{pos} \in \mathbb{R}^{ (N+1) \times D }   $ is the position embedding.

The resultant feature sequence $\mathbf{z}_0  \in \mathbb{R}^{ (N+1) \times D} $ serves as the input of Transformer encoder blocks~\cite{dosovitskiy2020image},
which are mainly composed of Multi-head Self-Attention (MSA), Layer Normalization (LN), Multilayer Perceptron (MLP) and residual connection as in Fig.\ref{fig:overview}. The Transformer encoder block is formulated as:
\begin{equation}
\begin{split}
&\mathbf{z}_{l}^{\prime}=\text{MSA} (LN(\mathbf{z}_{l-1}))+ \mathbf{z}_{l-1}   \\
&\mathbf{z}_{l}=\text{MLP} (LN(\mathbf{z}_{l}^{\prime}))+ \mathbf{z}_{l}^{\prime}.
\end{split}
\end{equation}
Here, $l$ is the depth of Transformer block and  $ l=1 \ldots L $.
The MLP consists of two linear layers with GELU activation.
Finally, the output embedding $ \mathbf{z}_{L} \in \mathbb{R}^{ (N+1) \times D }$ of Transformer is utilized for subsequent text recognition.

\subsection{Adaptive Addressing and Aggregation Modules} 
\label{sec:a3modules}

Traditional Vision Transformers~\cite{dosovitskiy2020image,deit} usually append a learnable $\mathbf{x}_{class}$ token to the sequence of patch embeddings, which directly collects and aggregates the meaningful information and serves as the image representation for the classification of the whole image.
While the task of scene text recognition aims to produce a sequence of character predictions, where each character is only related to a small patch of the image.
Thus, the global image representation $ \mathbf{z}_{L}^0 \in \mathbb{R}^{ D } $  is inadequate for the text recognizing task.
ViTSTR~\cite{ViTSTR} directly employs the first $T$ tokens of  $ \mathbf{z}_{L} $ for text recognition, where  $T$ is the maximum text length. Unfortunately, the rest tokens of $ \mathbf{z}_{L} $ are not fully utilized.

\begin{figure*}[t]\centering
 \includegraphics[width=0.85\textwidth]{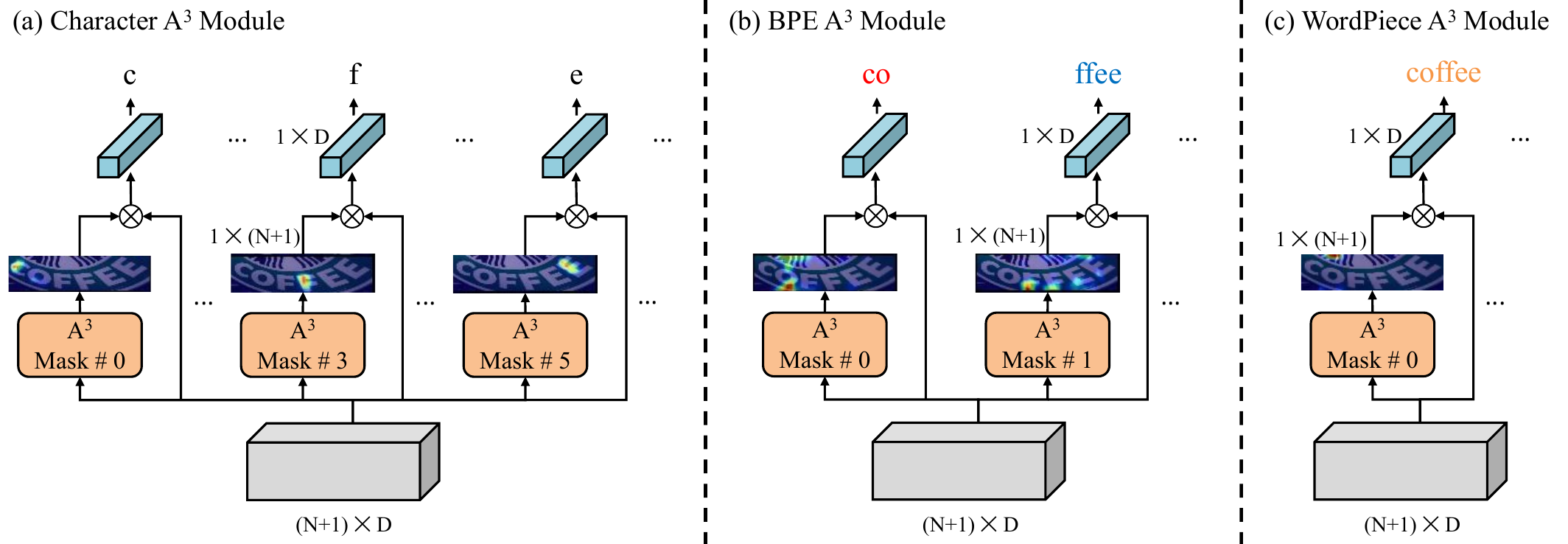}
 \caption{The detailed architectures of the three A$^3$ modules. }
 \label{fig:token}
\vspace{-3mm}
\end{figure*}

In order to take full advantage of the rich information of the sequence $ \mathbf{z}_{L} $ for text sequence prediction, 
we propose a tailored Adaptive Addressing and Aggregation (A$^3$)  module to select a  meaningful combination of tokens $ \mathbf{z}_{L} $ and integrate them into one token corresponding to a specific character.
Specifically, we manage to learn $T$ tokens $\mathbf{Y} = [\mathbf{y}_i]_{i=1}^{T}$ from the sequence $ \mathbf{z}_{L} $ for the subsequent text recognizing task.
An aggregation function is, thus, formulated as $\mathbf{y}_i = A_i(\mathbf{z}_{L})$, which converts the input  $ \mathbf{z}_{L} $ to a token vector $\mathbf{y}_i: \mathbb{R}^{ (N+1) \times D}  \mapsto  \mathbb{R}^{ 1 \times D } $. And  such $ T $ functions are constructed for the sequential output of text recognition.
Typically, the aggregation function $A_i(\mathbf{z}_{L})$ is implemented via a spatial attention mechanism~\cite{tokenlearner} to adaptively select the tokens from $ \mathbf{z}_{L} $ corresponding  to $i_{th}$ character.
Here, we employ function $\alpha_i(\mathbf{z}_{L})$ and softmax function to generate precise spatial attention 
mask $\mathbf{m}_i \in  \mathbb{R}^{ (N+1) \times 1}  $ from $\mathbf{z}_{L} \in \mathbb{R}^{ (N+1) \times D}$.
Thus, each output token $\mathbf{y}_i $ of A$^3$ module is produced by 
\begin{equation}
\begin{split}
&\mathbf{z}_{L} = LN(\mathbf{z}_{L}) \\
&\mathbf{y}_{i}= A_i(\mathbf{z}_{L}) = \mathbf{m}_i^T \tilde{\mathbf{z}}_{L}  = \text{softmax}(\alpha_i(\mathbf{z}_{L}))^T (\mathbf{z}_{L}\mathbf{U}) \\
&\mathbf{y}_{i} = LN(\mathbf{y}_{i}). \\
\end{split}
\end{equation}
Here, $\alpha_i$(·) is implemented by group convolution with one $1 \times 1$ kernel and $A_i$ represents the $i_{th}$ Addressing function to generate attention mask $\mathbf{m}_i \in  \mathbb{R}^{ (N+1) \times 1}$. And $ \mathbf{U}\in \mathbb{R}^{ D \times D}$  is a linear mapping matrix for learning feature $ \tilde{\mathbf{z}}_{L}$. Therefore, the resulting tokens of different aggregation functions are gathered together to form the final output tensor as follows:
\begin{equation}
\label{eq:Y}
\begin{split}
&\mathbf{Y}= [\mathbf{y}_{1}\mathbf{y}_{2};\ldots;\mathbf{y}_{T}] = [A_1(\mathbf{z}_{L}); A_2(\mathbf{z}_{L}); \ldots ;A_T(\mathbf{z}_{L}) ].\\
\end{split}
\end{equation}

Owing to the effective and efficient  A$^3$ module, the ultimate
representation of the text sequence is denoted as $\mathbf{Y} \in \mathbb{R}^{ T \times D}$ in Eq.~(\ref{eq:Y}). Then, a character classification head is built by $ \mathbf{G} = \mathbf{YW}^T \in \mathbb{R}^{ T \times K} $ for text recognition, where $\mathbf{W} \in \mathbb{R}^{ K \times D}$ is a linear mapping matrix, $K$ is the number of categories and $ \mathbf{G} $ is the classification logits. 
We regard this as the Character A$^3$ module for character-level prediction, of which the detailed structure is illustrated in Fig.~\ref{fig:token} (a). 

\subsection{Multi-Granularity Predictions}
\label{sec:MGP}
Character tokenization that simply splits text into characters is commonly used in scene text recognition methods.
However, this naive and standard way ignores the language information of the text. 
In order to effectively resort to linguistic information for scene text recognition, we incorporate subword~\cite{subword} tokenization mechanism in NLP~\cite{BERT} into the text recognition method. Subword tokenization algorithms aim to decompose rare words into meaningful subwords and remain frequently used words, so that the grammatical information of word has already been captured in the subwords. Meanwhile, since A$^3$ module is independent of Transformer encoder backbone, we can directly add extra parallel subword A$^3$ modules for subword predictions. 
In such a way, the language information can be implicitly injected into model learning for better performance. Notably, previous methods, \ie, SRN~\cite{SRN} and ABINet~\cite{ABInet}, design an explicit transformer module for language modeling, while we cast linguistic information encoding problem as a character and subword prediction task without an explicit language model. 

Specifically, we employ two subword tokenization algorithms Byte-Pair Encoding (BPE)~\cite{BPE} and WordPiece~\cite{wordpiece}
\footnote{Considering the potential out-of-vocabulary (OOV) issue in the inference phase, we did not directly predict whole words.} 
to produce various combinations, as shown in Fig.~\ref{fig:token}. 
Thus, the BPE A$^3$ module and WordPiece A$^3$ module are proposed for subword attention.
And two subword-level classification heads are used for subword predictions.
Since subwords could be whole words (such as ``coffee'' in WordPiece), subword-level and even word-level predictions can be generated by the BPE and WordPiece classification heads. 
Along with the original character-level prediction, we denote these various outputs as multi-granularity predictions for text recognition. In this way, character-level prediction guarantees the fundamental recognition accuracy, and subword-level or word-level predictions can serve as complementary results for noised images via linguistic information. 

Technically, the architecture of BPE or WordPiece A$^3$ module is the same as Character one. They are independent of each other with different parameters.
And the numbers of categories are different for different classification heads, which depend on the vocabulary size of each tokenization method.
The cross-entropy loss is employed for classification.
Additionally, the mask $\mathbf{m}_{i}$ precisely indicates the attention location of the $i_{th}$ character in Character A$^3$ module, 
while it roughly shows the $i_{th}$ subword region of the image in subword A$^3$ modules, due to the higher complexity and uncertainty of learning subword splitting. 

\begin{figure*}[t]\centering
 \includegraphics[width=0.9\textwidth]{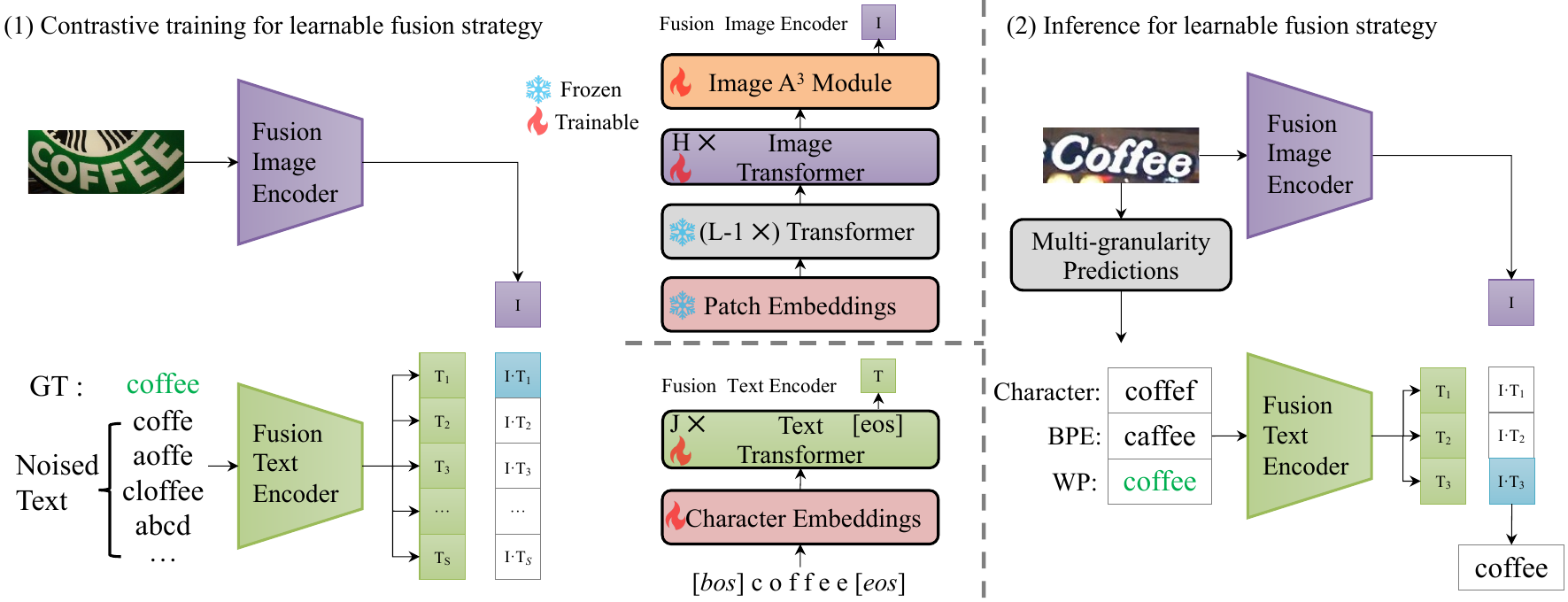}
 \caption{Schematic illustration of the proposed Learnable Fusion Strategy (LFS).}
 \label{fig:fusion}
\vspace{-2mm}
\end{figure*}

\subsection{Confidence-based Fusion Strategy}
\label{sec:fuse}
Multi-granularity predictions (Character, BPE and WordPiece) are generated by different A$^3$ modules and classification heads. Thus, a fusion strategy is required to merge these results. At the beginning, we attempt to fuse multi-granularity information by aggregating text features  $\mathbf{Y}$ of the output of different A$^3$ modules at the feature level. However, since these features are from different granularities, the $i_{th}$  token $\mathbf{y}_i^{char}$ of character level is not aligned with the $i_{th}$ token $\mathbf{y}_i^{bpe}$ (or $\mathbf{y}_i^{wp}$) of BPE level (or WordPiece level), so that these features cannot be directly added for fusion. Meanwhile, even if we concatenate features by [$\mathbf{Y}^{char}, \mathbf{Y}^{bpe}, \mathbf{Y}^{wp}$], only one character-level head can be used for final prediction. The subword information will be greatly impaired in this way, resulting in less improvement.

Instead, a decision-level fusion strategy is employed in our method. However, perfectly fusing these predictions is still a challenging problem~\cite{rrlrgu}.
We, therefore, propose a compromised but efficient Confidence-based Fusion Strategy (CFS) based on the prediction confidences of the three branches.
Specifically, the recognition confidence of each character or subword branch can be obtained by the corresponding classification head.
Then, we use two score functions $f(\cdot)$ to produce the final recognition score based on the atomic confidences:
\begin{equation}
\label{eq:mean}
f_{Mean}([c_1, c_2, \ldots, c_{eos}])= \frac{1}{n} \sum_{i=1}^{eos} c_i, \\
\end{equation}
\begin{equation}
\label{eq:prod}
f_{Cumprod}([c_1, c_2, \ldots, c_{eos}])= \prod_{i=1}^{eos}\ c_i.
\end{equation}

We only consider the confidence of valid characters or subwords and ending symbol $eos$, and ignore padding symbol $pad$.
``Mean'' recognition score is generated by the mean value function as in Eq.~(\ref{eq:mean}), while ``Cumprod'' represents the score produced by the cumulative product function.
Then, three recognition scores of three classification heads for one image can be obtained by $f(\cdot)$.
CFS simply picks the one with the highest fusion score as the final predicted result.

\subsection{Learnable Fusion Strategy} \label{sec:learnable}

Confidence-based Fusion Strategy (CFS) is both efficient and effective. However, since the raw confidence scores are produced from three different classification tasks at different levels, they are theoretically incomparable among each other, potentially limiting the effectiveness of CFS. Meanwhile, the ``Mean'' function might reduce the impact of individual character errors, and the ``Cumprod'' function may be skewed towards shorter character sequences. A predefined fusion function may not fully reflect the recognition quality of multi-granularity results. Therefore, we propose a Learnable Fusion Strategy (LFS) to directly measure the similarities between the predicted words and the original images, rather than relying on the raw confidence scores outputted by the three classification heads.

The main idea and module architecture of the proposed LFS are depicted in Fig.~\ref{fig:fusion}, where the image embedding $\mathbf{I} \in \mathbb{R}^{ 1 \times D} $ of the whole image is generated by the fusion image encoder
and the text embeddings $\mathbf{T}  \in \mathbb{R}^{ 1 \times D}$ are produced by the fusion text encoder.
Drawing inspiration from CLIP~\cite{CLIP}, we use contrastive learning to learn the alignment of text-image pairs.
The similarity scores between the image and three text sequences (words) produced by the three branches (Character, BPE and WordPiece) can be used for fusing the multi-granularity predictions. 
Noted that the ViT backbone and the three A$^3$ modules are frozen for generating the original multi-granularity predictions, while the new LFS module will be solely trained to fuse these results.
The details of the LFS module will be elaborated in the following subsections.

\subsubsection{Fusion Image Encoder}
The image embeddings are mainly derived from the ViT backbone as mentioned in Sec.~\ref{Sec:ViT}.
Specifically, we attach a trainable ``Adapter'' consisting of $H$ Transformer encoder blocks and a new Image A$^3$ module shown in Fig.~\ref{fig:fusion} to the $(L-1)_{th}$ layer of ViT backbone as in Fig.~\ref{fig:overview} for the whole image representation, denoted as $\mathbf{I}= A_I(\mathbf{\hat{z}})$,
where $\mathbf{\hat{z}}$ is produced by the output embeddings $\mathbf{{z}}_{L-1}$ of the $(L-1)_{th}$ frozen layers of ViT backbone within extra trainable Transformer encoder blocks.
Here, we do not directly utilize the outputs of the last layer $\mathbf{z}_{L}$ for image embedding extraction,
since $\mathbf{z}_{L}$ is customized for multi-granularity text recognition.
We deem that the outputs of the penultimate block $\mathbf{z}_{L-1}$ are more general features for computing the whole image representation.
Different from the Character A$^3$ module that produces $T$ tokens $\mathbf{Y} \in \mathbb{R}^{ T \times D}$ in Sec.\ref{sec:a3modules}, the Image A$^3$ module $A_I$ generates only one token $\mathbf{I} \in \mathbb{R}^{ 1 \times D}$ to represent the whole image.
Finally, the image feature is layer normalized and linearly projected into the multi-modal embedding space for contrastive learning.

\subsubsection{Fusion Text Encoder}

We adopt a text encoder with $J$ Transformer blocks to transcribe words into text embeddings based on character level tokenization.
All the characters in the words are modeled uniformly, meaning that various words including frequently used words, combinations of characters and numbers, or even random strings are encoded consistently and treated equally, avoiding potential 
imbalance issues.
Specifically, the text encoder is a Transformer with the architecture of the one in CLIP~\cite{gpt2,CLIP}.
The max input sequence length is $T$ and causal attention mask is used.
The character sequence of text is bracketed with $bos$ and $eos$ tokens and the activations of the last layer of the transformer at the $eos$
token are treated as the text feature representation $\mathbf{T} \in \mathbb{R}^{ 1 \times D}$.
Similar to the image features, layer normalization and linear projection are utilized.

\subsubsection{Training}
Given a batch of $M$ (\textit{text}, \textit{image}) pairs, CLIP aims to learn a multi-modal metric space by maximizing the similarities of the image and text embeddings of the $M$ true pairs in the batch, while minimizing the similarities of the
embeddings of the rest $M^2 - M$ incorrect pairings.
However, the target of LFS is to distinguish the subtle differences between characters within similar words referring to the image as shown in Fig.~\ref{fig:fusion} (1).
Thus, given a (\textit{text}, \textit{image}) pair, in the context of LFS, the negative text is constructed with \textit{\textbf{intentionally noised text variations}}, rather than the text from other pairs.
Concretely, we define $4$ single-character perturbations to delete, replace, repeat or insert at every character position in a word.
And we also design multi-character perturbations where we first randomly select $2$ or $3$ character positions, and then perform one random single-character perturbation at each selected position.
Notably, the noised text variations must be different from the ground-truth text.
Consequently, $N_s$ noised text is produced for one reference image.
Next, the contrastive representation learning with InfoNCE loss~\cite{infoNCE} is adapted for our image and text encoder learning,
which is computed as follows:
\begin{equation}
\begin{split}
& p_{\theta}( \mathbf{I}_i, \mathbf{T}_{i} )=\dfrac{\exp( S(\mathbf{I}_{i}, \mathbf{T}_{i}^{*}))}{\exp(S(\mathbf{I}_{i}, \mathbf{T}_{i}^{*})) + \sum_{k\in{{N}_{s}}}\exp(S(\mathbf{I}_{i},\mathbf{T}_{i}^{k}))}  \\
& L_{fuse} (\theta) = -\dfrac{1}{N_I} \sum_{i=1} \log \; p_{\theta}( \mathbf{I}_{i}, \mathbf{T}_{i}). \\
\end{split}
\end{equation}

Here, $S(\mathbf{I},\mathbf{T})$ represents the cosine similarity between image and text embeddings.
$\mathbf{T}_{i}^{*}$ is the embedding of the ground-truth text and $\mathbf{T}_{i}^{k}$ is the one of the noised text for the $i_{th}$ image.
${N_I}$ is the number of images.
$\theta$ represents the trainable parameters of LFS as illustrated in Fig.~\ref{fig:fusion} (1).

\subsubsection{Inference}

The inference process of LFS is shown in Fig.~\ref{fig:fusion}(2).
Given an input image $\mathbf{x}$, multi-granularity predictions are firstly produced by the A$^3$ modules as mentioned in Sec.~\ref{sec:MGP}. 
Then, the text embeddings of these $3$ predictions are generated by the fusion text encoder and the image embedding is produced by the fusion image encoder. 
Lastly, the cosine similarities between the image embedding and the $3$ text embeddings are computed, and the text with the highest similarity score will be chosen as the final result.

\begin{table*}[t]\centering
\setlength{\tabcolsep}{5pt}
\ra{1.2}
\caption{The ablation study of the proposed vision STR model and the accuracy comparisons with previous SOTA STR methods based on only vision information.}
\label{tab:char}
\begin{tabular}{|l|c|c|c|c|c|c|c|c|c|c|}
\hline
\textbf{Methods} & \textbf{Backbone} & \textbf{Image Size (Patch)} &\textbf{IC13} &\textbf{SVT}  &\textbf{IIIT}   & \textbf{IC15} & \textbf{SVTP} &\textbf{CUTE}  &\textbf{AVG} \\
\hline
MASTER~\cite{MASTER} &   \multirow{3}{*}{CNN}  & - &95.3 &90.60 &95.0 &79.4 &84.5 &87.5 &89.5	 \\
SRN$_V$~\cite{SRN}  &  & - &93.2 &88.1 &92.3 &77.5 &79.4 &84.7 &86.9	 \\
ABINet$_V$~\cite{ABInet}   &  & - &94.9 &90.4 &94.6 &81.7 &84.2 &86.5 &89.8 \\
\hline
MGP-STR$_{P=16}$ &  \multirow{3}{*}{ViT}   & $ 224 \times 224 (16 \times 16) $ &95.68	&91.96	&95.13	&83.88	&85.74	&90.28	&91.07	 \\
MGP-STR$_{P=4}$ &  & $ 32 \times 128 (4 \times 4) $ &96.62	&92.27	&95.40	&84.76	&86.98	&88.54	&91.58	 \\
MGP-STR$_{Vision}$  &  & $  32 \times 128 (4 \times 4)$ &96.50 &93.20 &{96.37} &86.25 &89.46 &{90.63} &92.73 \\
\hline
\end{tabular}
\end{table*}

\section{Experiment} \label{Sec:Experiment}

In this section, we present qualitative examples, quantitative comparisons, ablation studies and 
in-depth analyses, to prove the effectiveness and advantages of the proposed MGP-STR method.

\subsection{Datasets}
\label{Sec:data}
For fair comparisons, we use MJSynth~\cite{MJ1,MJ2} and SynthText~\cite{ST} as the training data. MJSynth contains $9M$ synthetic text images and SynthText includes $7M$ synthetic text images, respectively. The test data consists of ``regular'' and ``irregular'' datasets. 
The ``regular'' dataset is mainly composed of horizontally aligned text images. 
IIIT 5K-word (IIIT)~\cite{IIIT} consists of 3,000 images collected on the website. 
Street View Text (SVT)~\cite{SVT} contains 647 test images. 
ICDAR 2013 (IC13)~\cite{IC13} includes 1,095 images cropped from mall pictures, and has two versions for evaluation (\ie, IC13(857) and IC13(1015)).
The text instances in the ``irregular'' dataset are mostly curved or distorted. 
ICDAR 2015 (IC15)~\cite{IC15} includes 2,077 images collected from Google Eyes, and also has two versions for evaluation (\ie, IC15(1811) and IC15(2077)).
Street View Text-Perspective (SVTP)~\cite{SVTP} has 639 images collected from Google Street View. 
CUTE80 (CUTE)~\cite{CUTE} consists of 288 images with curved text.
Typically, IC13 and IC15 represent IC13(857) and IC15(1811) in the following experiments, unless otherwise stated.

We also study the performance of  MGP-STR trained on real data. Following~\cite{PARSeq}, we use COCO-Text (COCO)~\cite{COCO}, RCTW17~\cite{RCTW}, Uber-Text (Uber)~\cite{Uber}, ArT~\cite{ArT}, LSVT~\cite{LSVT}, MLT19~\cite{MLT}, ReCTS~\cite{ReCTS}, TextOCR~\cite{TextOCR} and OpenVINO~\cite{OpenVINO} as real training datasets. Please refer to \cite{realdata,PARSeq} for the comprehensive description of the real data.
Meanwhile, we also evaluate MGP-STR on the test sets of $3$ challenging datasets used in~\cite{PARSeq} for a more comprehensive comparison.
Specifically, COCO-Text (COCO)~\cite{COCO} includes 9,825 samples with low-resolution and occluded text.
ArT~\cite{ArT} contains 35,149 images with curved and rotated text.
Uber-Text (Uber)~\cite{Uber} has 80,418 images with vertical and rotated text instances.

To validate the universality of the proposed MGP-STR algorithm, we also conduct experiments on handwritten text datasets with English (IAM~\cite{IAM} and CVL~\cite{CVL}) and French (RIMES~\cite{RIMES}). In accordance with the methodologies employed in DiG~\cite{DiG}, we gather a training dataset of 146,805 images sourced from the training sets of CVL and IAM, and two test sets of 12,012 images and 13,752 images are assigned to CVL and IAM, respectively. For the RIMES dataset, we gather 51,737 images for training and 7,776 images for testing.



\subsection{Implementation Details}

\subsubsection{Model Configuration} 

In the default setting, MGP-STR is built upon the ViT model~\cite{dosovitskiy2020image}, which is composed of $12$ stacked Transformer blocks.
For each layer, the number of heads is $12$ and the embedding dimension $D$ is $768$. 
Notably, square $ 224 \times 224$ images~\cite{dosovitskiy2020image,deit,ViTSTR} are not adopted in our method.
The standard height $H$ and width $W$ of the input images are set to $32$ and $128$.
The patch size $P$ is set to 4 and thus $N=8 \times 32 =256$ plus one $[class]$ tokens $\mathbf{z}_L \in \mathbb{R}^{257\times768}$ will be produced.
The maximum length $T$ of the output sequence $\mathbf{Y}$ of A$^3$ module is set to $27$. 
The vocabulary size $K$ of the Character classification head is set to 38, including $0-9$, $a-z$, $pad$ for the padding symbol and $eos$ for the ending symbol.
The vocabulary sizes of BPE and WordPiece heads are $50,257$ and $30,522$, respectively.
In particular, since RIMES includes special characters of French, the vocabulary size $K$ is set to $50$, and we use a special WordPiece tokenization that supports French, which has $28,996$ subwords.
The evaluation on the handwritten text dataset is case-insensitive and excludes punctuations.

For the LFS module, an extra $H=1$ Transformer block is attached to the $11_{th}$ layer of ViT backbone as mentioned above for the image representation in fusion image encoder.
$J=2$ Transformer blocks are used for text feature extraction in the Fusion Text Encoder.
The maximum input text length $T$ is set to $27$.
$L2$ normalization is performed before the computation of cosine similarity.
$N_s =256$ noised text variations are generated by repeated $4$ single-character perturbations within $T / L_w * 5$ times and about $150$ multi-character perturbations, where $L_w$ is the length of the ground-truth text.


\subsubsection{Model Training} 

In the default setting, the pre-trained weights of the DeiT-base~\cite{deit} model are loaded as the initial weights for the ViT backbone of MGP-STR, except for the patch embedding module, due to inconsistent patch sizes.
Common data augmentation methods~\cite{randaug} for text images, such as perspective distortion, affine distortion, blur, noise and rotation, are applied.
We use $4$ NVIDIA A100 GPUs to train our model with a batch size of $100$. Adadelta~\cite{Adadelta} optimizer is employed with an initial learning rate of $1$.
The learning rate decay strategy is Cosine Annealing LR~\cite{cosinlr} and the training lasts about $100$ million iterations.

The trainable weights of LFS are randomly initialized.
LFS is trained with a batch size of $100$ on $2$ NVIDIA A100 GPUs. 
Adadelta~\cite{Adadelta} optimizer is used with an initial learning rate of $0.5$.
Besides the common data augmentation mentioned above, random earasing~\cite{erasing} data augmentation is introduced and $80\%$ images will be augmented.
Cosine Annealing LR is also used for learning rate decay and the training lasts about $200,000$ iterations.

\begin{table}[t]\centering
\setlength{\tabcolsep}{4pt}
\ra{1.2}
\caption{The accuracies of MGP-STR$_{CFS}$ with different confidence-based fusion strategies}
\label{tab:fuse}
\begin{tabular}{|l|c|c|c|c|c|c|c|c|}
\hline
\textbf{Mode} &\textbf{IC13} &\textbf{SVT}  &\textbf{IIIT}   & \textbf{IC15} & \textbf{SVTP} &\textbf{CUTE}  &\textbf{AVG} \\
\hline
Char &96.49	&93.66	&96.1	&86.14	&88.83	&89.58	&92.53	 \\
\hline
Mean &97.31	&94.28	&96.60	&86.97	&90.23	&90.97	&93.28	 \\
Cumprod &97.32	&94.74	&96.40	&87.24	&91.01	&90.28	&93.35	 \\
\hline
\end{tabular}
\vspace{-3mm}
\end{table}

\begin{table*}[t]\centering
\setlength{\tabcolsep}{5pt}
\ra{1.2}
\caption{The results of the four variants of MGP-STR model. ``Char'', ``BPE'' and ``WP'' at the ``Output'' column represent predictions of Character, BPE and WordPiece classification head  in each model, respectively. ``CFS'' represents the fused results via confidence-based fusion strategies.}
\label{tab:CBW}
\begin{tabular}{|l|c|c|c|c|c|c|c|c|c|c|c|c|}
\hline
\textbf{Methods} & \textbf{Char} & \textbf{BPE} & \textbf{WP} &\textbf{Output} &\textbf{IC13} &\textbf{SVT}  &\textbf{IIIT}   & \textbf{IC15} & \textbf{SVTP} &\textbf{CUTE}  &\textbf{AVG} \\
\hline
MGP-STR$_{Vision}$ &\checkmark & $ \times $ & $ \times $ &  Char &96.50 &93.20 &{96.37} &86.25 &89.46 &{90.63} &92.73 \\
\hline
\multirow{3}{*}{MGP-STR$_{C+B}$} &  \multirow{3}{*}{\checkmark} & \multirow{3}{*}{ \checkmark} &  \multirow{3}{*}{$ \times $} &  Char &97.43	&93.82	&96.53	&85.92	&89.15	&90.28	&92.84	 \\
& & & &  BPE &97.78	&94.13	&90.00	&81.12	&88.37	&82.64	&88.63	 \\
& & & &  CFS &97.67 &94.47	&96.73	&86.97	&88.99	&89.93	&93.24	 \\
\hline
\multirow{3}{*}{MGP-STR$_{C+W}$} &  \multirow{3}{*}{\checkmark} &  \multirow{3}{*}{$ \times $} & \multirow{3}{*}{ \checkmark}  &  Char &96.97	&93.97	&96.30	&86.20	&90.39	&89.93	&92.87	 \\
& & & &  WP &95.92	&93.35	&87.70	&78.74	&89.30	&80.21	&86.78	 \\
& & & &  CFS &97.32	&93.82	&96.60	&86.91	&90.54	&90.97	&93.25	 \\
\hline
\multirow{4}{*}{MGP-STR$_{CFS}$} &  \multirow{4}{*}{\checkmark} &  \multirow{4}{*}{\checkmark} & \multirow{4}{*}{ \checkmark}  &  Char &96.49	&93.66	&96.10	&86.14	&88.83	&89.58	&92.53	 \\
& & & &  BPE &95.56	&93.66	&88.73	&79.84	&89.76	&83.33	&87.63	 \\
& & & &  WP &95.79	&94.59	&86.37	&77.36	&89.61	&79.86	&85.99	 \\
& & & &  CFS &97.32	&94.74	&96.40	&87.24	&91.01	&90.28	&93.35	 \\
\hline
\end{tabular}
\end{table*}

\begin{table}[t]\centering
\setlength{\tabcolsep}{2pt}
\ra{1.5}
\caption{The results of MGP-STR$_{Vision}$ equipped with BCN and MGP-STR$_{CFS}$. ``V'' represents the results of the pure vision output. ``V+L'' represents the results based on both vision and language parts. $*$ represents the upper bound of MGP-STR$_{CFS}$. }
\label{tab:BCN}
\begin{tabular}{|l|c|c|c|c|c|c|c|c|c|}
\hline
\textbf{Methods} & \textbf{Mode} &\textbf{IC13} &\textbf{SVT}  &\textbf{IIIT}   & \textbf{IC15} & \textbf{SVTP} &\textbf{CUTE}  &\textbf{AVG} \\
\hline
MGP-STR$_{Vision}$  &  V &96.97	&93.82	&95.90	&85.53	&89.15	&89.58	&92.40	 \\
+BCN & V+L &97.32	&95.36	&95.97	&86.69	&91.78	&89.93	&93.14	 \\
\hline
MGP-STR$_{CFS}$   & V+L &97.32	&94.74	&96.40	&87.24	&91.01	&90.28	&93.35	 \\
MGP-STR$_{CFS}^*$   & V+L &97.66	&96.29	&96.97	&89.06	&92.09	&92.01	&94.38	 \\
\hline
\end{tabular}
\vspace{-3mm}
\end{table}

\subsection{Discussions on ViT and A$^3$ Modules} 
We analyze the influence of the patch size of Vision Transformer and the effectiveness of A$^3$ module in the proposed MGP-STR method (shown in Tab.~\ref{tab:char}). MGP-STR$_{P=16}$ represents the model that simply uses the first $T$ tokens of $\mathbf{z}_L$ for text recognition as in ViTSTR~\cite{ViTSTR}, where the input image is reshaped to $224 \times 224$ and the patch size is set to $16 \times 16 $. In order to retain the significant information of the original text image, $32 \times 128 $ images with $4 \times 4 $ patches are employed in MGP-STR$_{P=4}$. MGP-STR$_{P=4}$ outperforms MGP-STR$_{P=16}$, which indicates that the standard image size of ViT~\cite{dosovitskiy2020image,deit} is incompatible with the task of text recognition. Thus, $32 \times 128 $ images with $4 \times 4 $ patches are used in MGP-STR.

When the Character A$^3$ module is introduced into MGP-STR, denoted as MGP-STR$_{Vision}$, the recognition performance will be further improved. MGP-STR$_{P=16}$ and MGP-STR$_{P=4}$ cannot fully learn and utilize all the tokens, while the Character A$^3$ module can adaptively aggregate features of the last layer, resulting in more sufficient learning and higher accuracy. Meanwhile, compared with previous SOTA text recognition methods with CNN feature extractors, the proposed MGP-STR$_{Vision}$ method achieves substantial performance improvement. 

\subsection{Discussions on Multi-Granularity Predictions }

\subsubsection{Effect of Confidence-based Fusion Strategy }

Since the subwords generated by subword tokenization methods carry statistical and even grammatical information, we directly employ subwords as the targets of our model to capture the linguistic prior implicitly. As described in Sec.~\ref{sec:a3modules}, two different subword tokenizations BPE and WordPiece are employed for complementary multi-granularity predictions. 
Besides the character prediction, we propose two confidence-based fusion strategies to further merge the results from the three branches (corresponding to Character, BPE and WordPiece), denoted as ``Mean’’ and ``Cumprod’’ as mentioned in Sec.~\ref{sec:fuse}. We denote this method that merges the three results as MGP-STR$_{CFS}$, and the accuracies of MGP-STR$_{CFS}$ with different CFS fusion strategies are listed in Tab.~\ref{tab:fuse}. Additionally, the first line ``Char'' in Tab.~\ref{tab:fuse} records the result of character classification head in MGP-STR$_{CFS}$. As can be seen, both ``Mean’’ and ``Cumprod’’ fusion strategies can significantly improve the final recognition accuracy over that of single character-level results. Due to the better performance of ``Cumprod’’ strategy, we employ it as the default choice of Confidence-based Fusion Strategy (CFS) in the following experiments.

\subsubsection{Effect of Subword Representations}
We evaluate four variants of the MGP-STR model to verify the effect of subword representations.
The performances of these four variants are elaborately reported in Tab.~\ref{tab:CBW}, including the fused results and the results of each single classification head. Specifically, MGP-STR$_{Vision}$ with only Character A$^3$ module has already obtained promising results. MGP-STR$_{C+B}$ and MGP-STR$_{C+W}$ incorporate Character A$^3$ module with BPE A$^3$ module and WordPiece A$^3$ module, respectively. No matter which subword tokenization is used alone, the accuracy of ``CFS’’ can exceed that of ``Char’’ in both MGP-STR$_{C+B}$ and MGP-STR$_{C+W}$ methods, respectively. Notably, the performance of the classification head of ``BPE’’ or ``WP’’ could be better than that of ``Char’’ on the SVP and SVTP datasets in the same model. These results show that subword predictions can 
boost text recognition performance by implicitly introducing language information. Thus, MGP-STR$_{CFS}$ with three A$^3$ modules can produce complementary multi-granularity predictions. By fusing these multi-granularity results, MGP-STR$_{CFS}$ obtains better performance than single ones.

\subsubsection{Comparison with Bidirectional Cloze Network}
\label{sec:bcn}
Bidirectional Cloze Network (BCN) is designed in ABINet~\cite{ABInet} for explicit language modeling, and it leads to favorable improvement over the pure vision model. We equip MGP-STR$_{Vision}$ with BCN as a competitor of MGP-STR$_{CFS}$ to verify the advantage of the Multi-Granularity Prediction strategy. Concretely, we first reduce the dimension $768$ of feature $\mathbf{Y}$ to $512$ for aligning with the output of BCN. Following the training settings in~\cite{ABInet}, the results of this hybrid model are reported in Tab.~\ref{tab:BCN}. Apparently, the accuracy of ``V+L'' is further improved over the pure vision prediction ``V'' in MGP-STR$_{Vision}$+BCN, and better than the original ABINet~\cite{ABInet}. However, the performance of MGP-STR$_{Vision}$+BCN is a little worse than that of MGP-STR$_{CFS}$.

In addition, we provide the upper bound of the performance of MGP-STR$_{CFS}$, denoted as MGP-STR$_{CFS}^*$ in Tab. 4. If one of the three predictions (``Char'', ``BPE'' and ``WP'') is right, the final prediction is considered to be correct. 
The highest score of MGP-STR$_{CFS}^*$ demonstrates the good potential of multi-granularity predictions.
Moreover, MGP-STR$_{CFS}$ only requires two new subword prediction heads, rather than the design of a specific and explicit language model as in~\cite{ABInet,SRN}. 
These experiments demonstrate the superiority of multi-granularity predictions over BCN.

\subsection{Discussions on ViT Backbones}
\label{sec:ViTBackbones}

\subsubsection{Results with Different Sizes of ViT} 
All of the proposed MGP-STR models mentioned earlier are based on DeiT-Base~\cite{deit}. We also introduce two smaller models, namely DeiT-Small and DeiT-Tiny as presented in~\cite{deit} to further evaluate the effectiveness of MGP-STR$_{CFS}$. Specifically, the embedding dimensions of DeiT-Small and DeiT-Tiny are reduced to $384$ and $192$, respectively. Tab.~\ref{tab:y0} records the results of each prediction head of MGP-STR$_{CFS}$ with different ViT backbones. In general, fusing multi-granularity predictions can still improve the performance of pure character-level prediction with different backbones, and bigger models can achieve higher performances with the same classification heads. More importantly, the results of ``Char’’ with DeiT-Small and even DeiT-Tiny have already surpassed the SOTA pure CNN-based vision models, referring to Tab.~\ref{tab:char}. Therefore, MGP-STR$_{Vision}$ with small or tiny ViT backbones is also a competitive vision model and multi-granularity predictions can also work well with ViT backbones of different sizes, showing the good adaptability of the proposed MGP-STR method.

\begin{table}[t]\centering
\setlength{\tabcolsep}{2.5pt}
\ra{1.2}
\caption{The results of MGP-STR$_{CFS}$ with different sizes of ViT backbones.}
\label{tab:y0}
\begin{tabular}{|l|c|c|c|c|c|c|c|c|c|}
\hline
\textbf{Backbone} & \textbf{Output} &\textbf{IC13} &\textbf{SVT}  &\textbf{IIIT}   & \textbf{IC15} & \textbf{SVTP} &\textbf{CUTE}  &\textbf{AVG} \\
\hline
\multirow{4}{*}{DeiT-Tiny}  &  Char &93.47	&90.57	&93.93	&82.94	&81.71	&84.38	&89.36	 \\
&  BPE &87.40	&84.39	&83.17	&73.72	&77.83	&71.53	&80.48	 \\
&  WP &53.79	&45.44	&60.07	&52.57	&42.79	&42.71	&53.92	 \\
&  CFS &94.05	&91.19	&94.30	&83.38	&83.57	&84.38	&89.91	 \\
\hline
\multirow{4}{*}{DeiT-Small}  &  Char &95.92	&91.04	&94.97	&84.59	&85.89	&86.81	&91.01	 \\
&  BPE &96.27	&93.35	&89.37	&79.74	&86.67	&82.29	&87.61	 \\
&  WP &75.50	&70.48	&74.70	&66.81	&68.06	&62.15	&71.36	 \\
&  CFS &96.38	&93.51	&95.30	&86.09	&87.29	&87.85	&91.96	 \\
\hline
\multirow{4}{*}{DeiT-Base}  &  Char &96.49	&93.66	&96.10	&86.14	&88.83	&89.58	&92.53	 \\
&  BPE &95.56	&93.66	&88.73	&79.84	&89.76	&83.33	&87.63	 \\
&  WP  &95.79	&94.59	&86.37	&77.36	&89.61	&79.86	&85.99	 \\
&  CFS &97.32	&94.74	&96.40	&87.24	&91.01	&90.28	&93.35	 \\
\hline
\end{tabular}
\end{table}

\subsubsection{Results with Different Initial Weights} 
Besides DeiT-Base, we also initialize MGP-STR with various recent pre-trained ViT backbone models (\ie, DINO~\cite{DINO}, DINOv2~\cite{DINOv2}, MAE~\cite{mae} CLIP~\cite{CLIP} and BLIP~\cite{BLIP}) to verify the effectiveness of our method. We train MGP-STR with more iterations ($200$ million) to fully exploit the ability of the models, and the results with different initial weights are listed in Tab.~\ref{tab:clip}.
With more training iterations, the result of MGP-STR$_{CFS}$ (Deit-Base) will be improved.
Additionally, the pre-training weights of both the pure image pre-trained ViTs (\ie, DINO, DINOv2 and MAE) and the image encoder ViTs of vision-language models (\ie{CLIP and BLIP}) are compatible with MGP-STR.
As can be observed from Tab.~\ref{tab:clip}, once sufficiently trained, the ViT backbones from different pre-trained models almost work equally well with MGP-STR.
In the subsequent experiments, the DeiT-Base model with $200$ million iterations is used, due to its good performance.

\subsection{Discussions on Learnable Fusion Strategy}

\subsubsection{Effect of Learnable Fusion Strategy}
We train the LFS module based on MGP-STR$^{\dagger}_{CFS}$,
and the results of the model with LFS are detailedly depicted in Tab.~\ref{tab:learnable}.
Referring to Tab.~\ref{tab:y0}, the results of ``BPE'' and ``WP'' in MGP-STR$^{\dagger}$ are improved by a large margin than MGP-STR. We argue that the difficulty of subword recognition is higher than that of character recognition.
Due to the sufficient training time, the performances of ``BPE'' and ``WP'' heads can be improved further, resulting in a better-fused accuracy in MGP-STR$^{\dagger}_{CFS}$.
Moreover, it is worth noting that the upper bound (as explained in Sec.~\ref{sec:bcn}) of MGP-STR$^{\dagger}$ is $94.67 \%$.
Compared with MGP-STR$^{\dagger}_{CFS}$ ($93.70 \%$), MGP-STR$^{\dagger}_{LFS}$ ($94.05 \%$) unleashes the potential of the multi-granularity predictions further and achieves substantial improvement ($0.3\%$).
These results have proved the advantage of LFS over CFS.

\begin{table}[t]\centering
\setlength{\tabcolsep}{2.5pt}
\ra{1.2}
\caption{The results of MGP-STR$_{CFS}$ with different initial weights. The models with $\dagger$ are trained with $200$ million iterations. ``V'' represents the image pre-training models. ``V+L'' represents the vision-language pre-training models.}
\label{tab:clip}
\begin{tabular}{|l|c|c|c|c|c|c|c|c|c|}
\hline
\textbf{Backbone} & \textbf{Mode} &\textbf{IC13} &\textbf{SVT}  &\textbf{IIIT}   & \textbf{IC15} & \textbf{SVTP} &\textbf{CUTE}  &\textbf{AVG} \\
\hline
DeiT & V  &97.32  &94.74 &96.40 &87.24 &91.01	&90.28 &93.35 \\
DeiT$^\dagger$ &  V &97.43	&95.52	&96.77	&87.80	&90.23	&91.32	&93.70	 \\
\hline
MAE$^\dagger$ & V &97.32 &95.36 & 96.43 &87.96 &90.54 & 91.32 & 93.60	 \\
DINO$^\dagger$ & V &97.78 &95.21	&96.27	&87.58	&90.70	&90.97 & 93.47	 \\
DINOv2$^\dagger$ & V &97.55 &95.05 &96.53 &87.63 &90.54 &92.36 & 93.59	 \\
\hline
CLIP$^\dagger$  & V+L &97.67 &94.59	&96.20	&87.91	&91.63	&92.36 & 93.60	 \\
BLIP$^\dagger$ & V+L &97.67 &95.05 &96.03 &87.85 &91.78 &91.32 	&93.53 \\
\hline
\end{tabular}
\end{table}

\begin{table}[t]\centering
\setlength{\tabcolsep}{3pt}
\ra{1.3}
\caption{The results of MGP-STR$^\dagger$ with ``CFS'', and ``LFS''. $\dagger$ represents $200$ million training iterations, $*$ represents the upper bound of MGP-STR$^{\dagger}$. }
\label{tab:learnable}
\begin{tabular}{|l|c|c|c|c|c|c|c|c|c|}
\hline
\textbf{Methods} &\textbf{IC13} &\textbf{SVT}  &\textbf{IIIT}   & \textbf{IC15} & \textbf{SVTP} &\textbf{CUTE}  &\textbf{AVG} \\
\hline
MGP-STR$^\dagger_{Char}$  &96.97	&93.97	&96.53	&86.53	&88.22	&89.93	&92.85	 \\
MGP-STR$^\dagger_{BPE}$  &96.97	&94.90	&90.27	&81.67	&89.77	&85.07	&89.07	 \\
MGP-STR$^\dagger_{WP}$  &97.20	&95.36	&89.23	&79.90	&91.63	&82.64	&88.34	 \\
\cline{1-9}
MGP-STR$^\dagger_{CFS}$  &97.43	&95.52	&96.77	&87.80	&90.23	&91.32	&93.70	 \\
MGP-STR$^\dagger_{LFS}$  &97.78	&96.29	&96.77	&88.13	&92.09	&91.32	&94.05	 \\
\cline{1-9}
MGP-STR$^\dagger_{CFS^*}$  &98.13 &96.45 &97.23  &89.29 &93.02  &91.67  &94.67   \\
\hline
\end{tabular}
\end{table}

\subsubsection{Effect of the Number of Layers} 
We construct $4$ variants of LFS with different numbers of layers, and the results are recorded in Tab.~\ref{tab:depth}.
Both fusion image encoder and fusion text encoder are based on Transformer layers, and the number of layers in Tab.~\ref{tab:depth} means the trainable layers as described in Sec.~\ref{sec:learnable}.
We can see that all the models, even model (a) with a single layer, can achieve similar performances.
Meanwhile, the latency of LFS is negligible (about $3$ ms).
We claim that the image information has already been encoded in the ViT backbone and $1$ more extra layer is enough for computing the whole image representation. 
And the word information with characters can be captured with $2$ transformer layers.
In summary, the proposed LFS module is an effective alternative for better fusion results and the costs of training and inference brought by LFS are quite limited.
Therefore, we chose model (c) as the default setting of the proposed LFS, due to the good result and high efficiency.

\begin{table}[t]\centering
\setlength{\tabcolsep}{3pt}
\ra{1.2}
\caption{The results of the four variants of the LFS module. ``Text'' and ``Image'' represent the number of layers of fusion text encoder and fusion image encoder, respectively. Additionally, the results include the inference time and model size of LFS.}
\label{tab:depth}
\begin{tabular}{|c|c|c|c|c|c|c|c|c|}
\hline
\textbf{\multirow{2}{*}{Model}} &\textbf{\multirow{2}{*}{Text}} &\textbf{\multirow{2}{*}{Image} }&\textbf{IC13} &\textbf{SVT}  &\textbf{IIIT}   & \textbf{\multirow{2}{*}{AVG}}  & \textbf{Time} &   \textbf{Params}\\
& & & \textbf{IC15} & \textbf{SVTP} &\textbf{CUTE}  & & \textbf{(ms)} &  \textbf{($ \times 10^6$)} \\
\hline
\multirow{2}{*}{(a)}  & \multirow{2}{*}{1} & \multirow{2}{*}{1} &98.02	&96.29	&96.57	&\multirow{2}{*}{94.03} & \multirow{2}{*}{2.48} & \multirow{2}{*}{15.6}  \\
& & &88.02	&92.56	&91.67	&	&	&	 \\
\hline
\multirow{2}{*}{(b)}  & \multirow{2}{*}{1} & \multirow{2}{*}{2} &97.90	&96.29	&96.50	&\multirow{2}{*}{93.97}	& \multirow{2}{*}{3.02}	& 	\multirow{2}{*}{22.6}  \\
& & &88.13	&92.09	&91.67	&	& &	 \\
\hline
\multirow{2}{*}{(c)}  & \multirow{2}{*}{2} & \multirow{2}{*}{1}  &97.78	&96.29	&96.77	&\multirow{2}{*}{94.05} 	& \multirow{2}{*}{3.05}	&  \multirow{2}{*}{22.6}	\\
& & &88.13	&92.09	&91.32	&	&	&	 \\
\hline
\multirow{2}{*}{(d)}  & \multirow{2}{*}{2} & \multirow{2}{*}{2} &98.02	&96.29	&96.57	&\multirow{2}{*}{94.00} 	& \multirow{2}{*}{3.65}	& \multirow{2}{*}{29.7} \\
& & &88.13	&92.09	&91.32	&	&	&	 \\
\hline
\end{tabular}
\vspace{-3mm}
\end{table}

\begin{table}[t]\centering
\setlength{\tabcolsep}{2.5pt}
\ra{1.2}
\caption{The results of MGP-STR$^{\dagger}_{LFS}$ trained with different numbers of noised text variants.}
\label{tab:neg}
\begin{tabular}{|c|c|c|c|c|c|c|c|c|c|}
\hline
\textbf{Models} &\textbf{Number} &\textbf{IC13} &\textbf{SVT}  &\textbf{IIIT}   & \textbf{IC15} & \textbf{SVTP} &\textbf{CUTE}  &\textbf{AVG} \\
\hline
CFS &  - &97.43	&95.52	&96.77	&87.80	&90.23	&91.32	&93.70	 \\
\hline
\multirow{4}{*}{LFS} &  8 &97.55	&95.67	&96.50	&87.74	&91.32	&90.63	&93.67 \\
&  64 &97.78	&95.98	&96.53	&87.85	&91.94	&90.63	&93.82	 \\
&  128 &97.90	&96.29	&96.57	&88.02	&92.25	&91.32	&93.97	 \\
&  256 &97.78	&96.29	&96.77	&88.13	&92.09	&91.32	&94.05	 \\
\hline
\end{tabular}
\vspace{-3mm}
\end{table}

\subsubsection{Effect of the Number of Noised Text} 
We construct experiments to investigate the effect of the number of negative pairs, which is essential in contrastive learning.
The results of models trained with $4$ different numbers of noised text are listed in Tab.~\ref{tab:neg}.
In general, the more noised text, the relatively higher the fused accuracies.
Specifically, the fusion text encoder can not be sufficiently learned with less noised text variants(\ie, $8$ or $64$).
Thus, MGP-STR$^{\dagger}_{LFS}$ with $8$ or $16$ noised text can not obtain obvious advantage over MGP-STR$^{\dagger}_{CFS}$.
However, when the number of noised text instances increases to $128$ or $256$, MGP-STR$^{\dagger}_{LFS}$ can achieve appealing improvements ($0.3 \%$) over MGP-STR$^{\dagger}_{CFS}$.
Due to the training efficiency, we set the number of noised text instances to $256$ in the proposed LFS module.

\subsection{Inside Details of MGP-STR}  
\subsubsection{Details of Multi-Granularity Predictions of CFS}  
\label{Sec:fa}

We show the detailed prediction process of the proposed MGP-STR$_{CFS}$ method on $6$ test images from standard datasets in Tab.~\ref{tab:fa}. In the first three images, the results of character-level prediction are incorrect, due to irregular font, motion blur and curved shape, respectively. The scores of character prediction are very low, since the images are difficult to recognize and one character is wrong in each image. However, ``BPE'' and ``WP'' heads can recognize ``table'' image with high scores. And ``BPE'' can make correct predictions with two subwords on ``dvisory'' and ``watercourse'' images, while ``WP''  is wrong for the ``watercourse'' image. After fusion, the mistakes can be corrected. From the rest three images, interesting phenomena can be observed. The predictions of ``Char'' and ``BPE''  conform to the images. The predictions of ``WP'', however, attempt to produce strings with more linguistic content, like ``today'' and ``guide''.
Generally, ``Char'' aims to produce characters one by one, while ``BPE'' usually generates n-gram segments related to images, and ``WP'' tends to directly predict whole words that are linguistically meaningful. These prove that the predictions of different granularities convey text information in different aspects and are indeed complementary.

\begin{table*}[t]\centering
\setlength{\tabcolsep}{5pt}
\ra{1.2}
\caption{The details of multi-granularity predictions of MGP-STR$_{CFS}$, including the CFS recognition scores (Score) of each prediction head, the intermediate multi-granularity (Gra.) results and the predictions (Pred.). (Best viewed in color.)}
\label{tab:fa}
\begin{tabular}{|c|c|c|c|c|c|c|c|}
\hline
\textbf{Images} & \textbf{GT} &\textbf{Output} &\textbf{Char}  &\textbf{BPE}   & \textbf{WP}  & \textbf{CFS} \\
\hline
    \multirow{3}{*}{
    \begin{minipage}[b]{0.15\columnwidth}
		\centering
		\raisebox{-.5\height}{\includegraphics[width=\linewidth]{CUTE80_tabbe_0.1643_table_0.9521_table_0.9813_table_0.9813_table.png}}
	\end{minipage}
	}
	& \multirow{3}{*}{table}
	&Score & 0.1643 & 0.9813 & 0.9521 & 0.9813\\
	\cline{3-7}
	& &Gra. & tabbe & \textcolor{orange}{table} & \textcolor{orange}{table}  & -\\
	\cline{3-7}
	& &Pred. & tabbe & table &  table & table \\
\hline
    \multirow{3}{*}{
    \begin{minipage}[b]{0.15\columnwidth}
		\centering
		\raisebox{-.5\height}{\includegraphics[width=\linewidth]{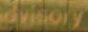}}
	\end{minipage}
	}
	& \multirow{3}{*}{dvisory}
	&Score &0.0316 & 0.8218 & 0.2574 & 0.8218\\
	\cline{3-7}
	& &Gra. &divsoory &  \textcolor{red}{d}  \textcolor{blue}{visory} &  \textcolor{orange}{dvisory} & - \\
	\cline{3-7}
	& &Pred. &divsoory & dvisory  & dvisory & dvisory \\
	\hline
    \multirow{3}{*}{
    \begin{minipage}[b]{0.15\columnwidth}
		\centering
		\raisebox{-.5\height}{\includegraphics[width=\linewidth]{SVT_watercourss_0.1565_waterco_0.632_water_course_0.8295_watercourse_0.8295_watercourse.png}}
	\end{minipage}
	}
	& \multirow{3}{*}{watercourse}
	&Score &0.1565 & 0.8295 & 0.632 & 0.8295\\
	\cline{3-7}
	& &Gra. & watercourss &  \textcolor{red}{water}  \textcolor{blue}{course} &  \textcolor{orange}{waterco} & - \\
	\cline{3-7}
	& &Pred. &watercourss & watercourse  & waterco & watercourse \\
		\hline
    \multirow{3}{*}{
    \begin{minipage}[b]{0.15\columnwidth}
		\centering
		\raisebox{-.5\height}{\includegraphics[width=\linewidth]{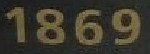}}
	\end{minipage}
	}
	& \multirow{3}{*}{1869}
	&Score &0.9999 & 0.9207 & 0.0354 & 0.9999\\
	\cline{3-7}
	& &Gra. & 1869 &  \textcolor{red}{18}  \textcolor{blue}{69} &  \textcolor{orange}{18} & - \\
	\cline{3-7}
	& &Pred. &1869 & 1869  & 18 & 1869 \\
	\hline
    \multirow{3}{*}{
    \begin{minipage}[b]{0.12\columnwidth}
		\centering
		\raisebox{-.5\height}{\includegraphics[width=\linewidth]{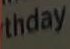}}
	\end{minipage}
	}
	& \multirow{3}{*}{thday}
	&Score &0.9998 & 0.5983 & 0.7638 & 0.9998 \\
	\cline{3-7}
	& &Gra. & thday &  \textcolor{red}{th}  \textcolor{blue}{day} &  \textcolor{orange}{today} & - \\
	\cline{3-7}
	& &Pred. &thday & thday  & today & thday \\
				\hline
    \multirow{3}{*}{
    \begin{minipage}[b]{0.12\columnwidth}
		\centering
		\raisebox{-.5\height}{\includegraphics[width=\linewidth]{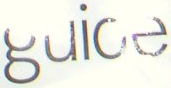}}
	\end{minipage}
	}
	& \multirow{3}{*}{guide}
	&Score &0.9675 & 0.6959 & 0.1131 & 0.9675 \\
	\cline{3-7}
	& &Gra. & guice &  \textcolor{red}{gu}  \textcolor{blue}{ice} &  \textcolor{orange}{guide} & - \\
	\cline{3-7}
	& &Pred. &guice & guice  & guide & guice \\
	\hline
\end{tabular}
\end{table*}

\begin{table*}[t]\centering
\setlength{\tabcolsep}{2.5pt}
\ra{1.2}
\caption{The details of MGP-STR$_{LFS}$, including the CFS recognition scores (Score) and the final prediction (Pred.) of each prediction head. The similarities between each type of prediction and input images given by LFS are listed at (Sim.). }
\label{tab:lfs}
\begin{tabular}{|c|c|c|c|c|c|c|c|}
\hline
\textbf{Images} & \textbf{GT} &\textbf{Output} &\textbf{Char}  &\textbf{BPE}   & \textbf{WP}  & \textbf{CFS} & \textbf{LFS} \\
\hline
    \multirow{3}{*}{
    \begin{minipage}[b]{0.15\columnwidth}
		\centering
		\raisebox{-.5\height}{\includegraphics[width=\linewidth]{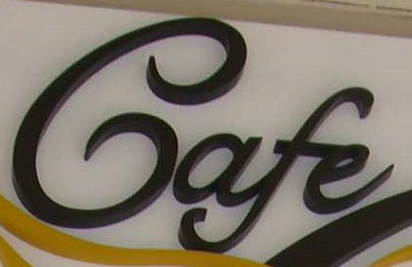}}
	\end{minipage}
    }
    & \multirow{3}{*}{cafe} &Score & 0.33 & 0.81 & 0.28 & 0.81 & - \\
    \cline{3-8}
    & & Sim. & 0.05 & 0.02 & 0.15 & - & 0.15 \\
    \cline{3-8}
    & & Pred. & caf & c & cafe & c & cafe \\
   \hline

    \multirow{3}{*}{
    \begin{minipage}[b]{0.23\columnwidth}
		\centering
		\raisebox{-.5\height}{\includegraphics[width=\linewidth]{SVT_200_HOTEL_horel_char88_hoel_bpe69_hotel_wp79_horel_fuse88_hotel_learn_28_17_33.jpg}}
	\end{minipage}
    }
    & \multirow{3}{*}{hotel} &Score & 0.88 & 0.69 & 0.79 & 0.88 & - \\
    \cline{3-8}
    & &Sim. & 0.28 & 0.17 & 0.33 & - & 0.33 \\
    \cline{3-8}
    & & Pred. & horel & hoel & hotel & horel & hotel \\
    \hline

    \multirow{3}{*}{
    \begin{minipage}[b]{0.15\columnwidth}
		\centering
		\raisebox{-.5\height}{\includegraphics[width=\linewidth]{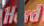}}
	\end{minipage}
    }
    & \multirow{3}{*}{hard} &Score & 0.45 & 0.26 & 0.44 & 0.45 & - \\
    \cline{3-8}
    & &Sim. & 0.07 & 0.07 & 0.08 & - & 0.08 \\
    \cline{3-8}
    & & Pred. & herd & herd & hard & herd & hard \\
    \hline

    \multirow{3}{*}{
    \begin{minipage}[b]{0.23\columnwidth}
		\centering
		\raisebox{-.5\height}{\includegraphics[width=\linewidth]{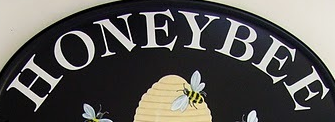}}
	\end{minipage}
    }
    & \multirow{3}{*}{honeybee} &Score & 0.75 & 0.95 & 0.33 & 0.95 & - \\
    \cline{3-8}
    & &Sim. & 0.11 & -0.02 & 0.21 & - & 0.21 \\
    \cline{3-8}
    & & Pred. & honeybec & honey & honeybee & honey & honeybee \\
    \hline

    \multirow{3}{*}{
    \begin{minipage}[b]{0.25\columnwidth}
		\centering
		\raisebox{-.5\height}{\includegraphics[width=\linewidth]{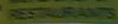}}
	\end{minipage}
    }
    & \multirow{3}{*}{restaurants} &Score & 0.01 & 0.35 & 0.07 & 0.35 & - \\
    \cline{3-8}
    & &Sim. & -0.24 & -0.06 & 0.03 & - & 0.03 \\
    \cline{3-8}
    & & Pred. & restuuannts & resturants & restaurants & resturants & restaurants \\
	\hline

    \multirow{3}{*}{
    \begin{minipage}[b]{0.28\columnwidth}
		\centering
		\raisebox{-.5\height}{\includegraphics[width=\linewidth]{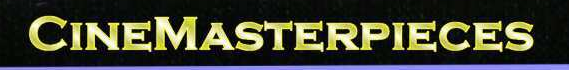}}
	\end{minipage}
    }
    & \multirow{3}{*}{cinemasterpieces} &Score & 0.90 & 0.85 & 0.07 & 0.90 & - \\
    \cline{3-8}
    & &Sim. & 0.33 & 0.38 & 0.23 & - & 0.38\\
    \cline{3-8}
    & & Pred. & cinemasterpiecss & cinemasterpieces & cinemaspiece & cinemasterpiecss & cinemasterpieces\\
    \hline
\end{tabular}
\end{table*}

\subsubsection{Details of Fusion Process of LFS}  \label{Sec:explfs}

We show several typical cases where the results of MGP-STR$_{CFS}$ are wrong, but MGP-STR$_{LFS}$ make correct predictions, to analyze the working mechanism of LFS.
The detailed predictions and similarities produced by MGP-STR$_{LFS}$ on $6$ test images are illustrated in Tab.\ref{tab:lfs}.
Generally, all ``Char'' predictions in Tab.\ref{tab:lfs} are wrong, due to irregular fonts, distortions or long text.
And the correct results of ``BPE'' and ``WP'' heads may be with lower CFS recognition scores, leading to wrong fusion results for MGP-STR$_{CFS}$.
In contrast, the similarities generated by LFS can be utilized to correct these errors.
Key observations from Tab.\ref{tab:lfs} are: 
(1) The similarities between the image and the wrong text with unmatched length are obviously lower than ones with matched length ( \eg, ``cafe'' $>$ ``caf''  $>$ ``c'', ``restaurants'' $>$ ``resturants''  $>$ ``restuuannts'',  and ``honeybee'' $>$ ``honeybec''  $>$ ``honey'').
(2) Even if only one character is incorrect, the similarity value will decrease drastically, \eg, $10\%$ of ``honeybee'' over ``honeybec'' and $5\%$ of ``cinemasterpieces'' over ``cinemasterpiecss''.
These qualitative analyses demonstrate that the similarity values of LFS can faithfully reflect the matching degrees between the images and the text sequences (even with subtle distinctions). In conclusion, LFS is a more effective way for fusing multi-granularity predictions than CFS.

\begin{table*}[t]
\setlength{\tabcolsep}{5pt}
\ra{1.3}
\centering
\caption{The comparisons with other STR methods on $8$ public standard benchmarks. The models with $\dagger$ are trained with $200$ million iterations. For training data, ``S'' represents synthetic training datasets (MJSynth and SynthText), S$^*$ means external text datasets are used for the language model, and ``R'' represents real datasets provided by~\cite{PARSeq}. The \textbf{bold} and \underline{underline} numbers represent the best and the second-best results, respectively. ``AVG'' is the average accuracy except those of IC13 (1015) and IC15 (2077).}
\label{tab:sota}
\begin{tabular}{|l|c|c|c|c|c|c|c|c|c|c|c|}
\hline
\textbf{\multirow{3}{*}{Methods}}  &\textbf{\multirow{3}{*}{Year}} &  &\multicolumn{4}{c|}{\textbf{Regular Text}} &\multicolumn{4}{c|}{\textbf{Irregular Text}} & \textbf{\multirow{2}{*}{AVG} }\\
\cline{4-11}
& &\textbf{Train} &\textbf{IC13} &\textbf{IC13} &\textbf{SVT}  &\textbf{IIIT}   & \textbf{IC15} & \textbf{IC15} & \textbf{SVTP} &\textbf{CUTE} &  \\
\cline{4-12}
& &\textbf{data} &\textbf{857} &\textbf{1015} &\textbf{647}  &\textbf{3000}   & \textbf{1811}  & \textbf{2077} & \textbf{645} &\textbf{288} & \textbf{7248} \\
\hline
TBRA~\cite{deep} &ICCV (2019) &S &93.6 &92.3 &87.5 &87.9 &77.6 & 71.8 &79.2  & 74.0  & 84.6 \\
ViTSTR-Base~\cite{ViTSTR} &ICDAR (2021) &S &93.2 &92.4 & 87.7 & 88.4 &78.5 &72.6 &81.8 &81.3 &85.6 \\
ESIR~\cite{ESIR} &CVPR (2019) &S  &-  &91.3 &90.2 &93.3 &- &76.9 &79.6 &83.3  &-  \\
DAN~\cite{DAN} &AAAI (2020) &S &-  &93.9 &89.2 &94.3  &- &74.5  &80.0 &84.4   & -  \\
SE-ASTER~\cite{SEED} &CVPR (2020) &S  &- &92.8 &89.6 &93.8 &80.0 & &81.4 &83.6  & 88.3  \\
TrOCR-Base~\cite{TrOCR} &AAAI (20223) &S &97.3	&96.3 &91.0	&90.1	&81.1 &75.0	&90.7 &86.8 & 88.7 \\
RobustScanner~\cite{RobustScanner}  &ECCV (2020) &S  &- &94.8 & 88.1  & 95.3 &- & 77.1 & 79.5 &  90.3 & - \\
TextScanner~\cite{TextScanner} &AAAI (2020) &S &- &92.9 &90.1 &93.9 &79.4 & &84.3 &83.3  & - \\
SATRN~\cite{SATRN} &CVPRW (2020)  &S  &- &94.1 &91.3 &92.8  &- &79.0 &86.5 &87.8 &- \\
MASTER~\cite{MASTER} &PR (2021) &S &-	&95.3 &90.6	&95.0	&- &79.4 &84.5	&87.5 & - \\
SRN~\cite{SRN} &CVPR (2020)  &S  & 95.5 & - & 91.5  & 94.8  &82.7 & - &85.1   &87.8 & 90.4 \\
VisionLAN~\cite{vlan}  &ICCV (2021) &S &95.7 & - &91.7& 95.8 &83.7 & - &86.0 &88.5 &91.2 \\
PARSeq~\cite{PARSeq} &ECCV (2022) &S &96.3	&95.5 &92.6	&95.7	&85.1 &81.4	&87.9	&91.4 & 92.0 \\
ABINet++~\cite{ABInetv2}  &TPAMI (2023) &S$^*$  &97.4 & 95.7 &93.5 &96.2 &86.0 &\textbf{85.1} &89.3 &89.2  & 92.6 \\
LevOCR~\cite{matrn} &ECCV (2022) &S$^*$ &96.9 &- &92.9 &96.6	&86.4 &-	&88.1	&91.7 & 92.8 \\
PTIE~\cite{ptie} &ECCV (2022) &S &-	&\textbf{97.2} &94.9	&96.3	&87.8 &\underline{84.3} &90.1	&\underline{91.7} &- \\
MATRN~\cite{matrn} &ECCV (2022) &S$^*$ &\textbf{97.9}	&95.8 &95.0	&96.6	&86.6 &82.8	&90.6	&\textbf{93.5} & 93.5 \\
MGP-STR$_{CFS}$~\cite{mgp} &ECCV (2022) &S  &{97.32} &96.55 &94.74 &96.40	&87.24 & 83.78	&\underline{91.01}	&90.28	&93.35 \\
\hline
MGP-STR$^{\dagger}_{CFS}$ &Ours &S &97.43	& 96.65 &\underline{95.52}	&\underline{96.77}	&\underline{87.80}	& 83.78 &90.23	&91.32	&\underline{93.70} \\
MGP-STR$^{\dagger}_{LFS}$ &Ours &S &\underline{97.78}	&\underline{96.95} &\textbf{96.29}	&\textbf{96.77}	&\textbf{88.13}	&84.11 &\textbf{92.09}	&91.32	&\textbf{94.05}	 \\
\hline
PARSeq~\cite{PARSeq} &ECCV (2022) &R &98.0	&98.1 &97.5	&98.3	&89.6 &88.4	&94.6	&97.7 &95.7 \\
ABINet++~\cite{ABInetv2}  &TPAMI (2023) &R  &98.0 &97.8 &\underline{97.8} &\textbf{98.6} &90.2 &88.5 &93.9 &97.7  &95.9 \\
MGP-STR$^{\dagger}_{CFS}$ &Ours &R  & \underline{98.37} & \underline{98.13} & 97.68  & 97.93  &\underline{90.89} & \underline{89.55} &\underline{95.35}   &\underline{97.92} & \underline{95.97} \\
MGP-STR$^{\dagger}_{LFS}$ &Ours &R  & \textbf{98.60} & \textbf{98.42} & \textbf{98.30}  &\underline{98.40}  &\textbf{91.06} & \textbf{89.79} &\textbf{96.59}   &\textbf{97.92} & \textbf{96.40} \\
\hline
\end{tabular}
\end{table*}

\subsection{Comparisons with Previous State-of-the-Arts} 
\label{Sec:sota}

\subsubsection{Results on Standard Benchmarks}

We compare the proposed MGP-STR$^{\dagger}_{CFS}$ and MGP-STR$^{\dagger}_{LFS}$ methods with previous state-of-the-art scene text recognition methods, and the results on $8$ standard benchmarks are summarized in Tab.~\ref{tab:sota}. All of the compared methods and ours are trained on synthetic datasets MJ and ST for a fair evaluation. And the results are obtained without any lexicon-based post-processing. Generally, language-augmented methods (\ie, SRN~\cite{SRN}, VisionLAN~\cite{vlan}, PARSeq~\cite{PARSeq}, ABINet++~\cite{ABInetv2}, LevOCR~\cite{levocr}, MATRN~\cite{matrn} and MGP-STR) perform better than language-free methods, showing the significance of linguistic information. PTIE~\cite{ptie}, which utilizes a transformer-only model with multiple patch resolutions, also achieves good results. 

Notably, owing to the multi-granularity predictions, MGP-STR$^{\dagger}_{CFS}$ has already outperformed the recent state-of-the-art method MATRN. Furthermore, MGP-STR$^{\dagger}_{LFS}$ obtains more impressive results by resorting to the proposed LFS over MGP-STR$^{\dagger}_{CFS}$.
Particularly, MGP-STR$^{\dagger}_{LFS}$ achieves the best results on $4$ out of $8$ benchmarks, outperforming the second best method MATRN by $1.3 \%$ on SVT, $0.17 \%$ on IIIT, $1.5 \%$ on IC15 (1811) and $2.3 \%$ on SVTP.
Moreover,  MGP-STR$^{\dagger}_{LFS}$ can attain the second best performance on IC13(857), IC13(1015) and IIIT. 
And MGP-STR$^{\dagger}_{LFS}$ also achieves comparable results on IC15(2077) and CUTE.
Consequently, MGP-STR$^{\dagger}_{LFS}$ achieves a substantial improvement of $0.7\%$ over MATRN in terms of average accuracy.

Following PARSeq~\cite{PARSeq}, we also train MGP-STR with real data to further study the potential of our method.
The models trained with real data as shown in Tab.~\ref{tab:sota} utilize the training and testing datasets provided by ~\cite{PARSeq} for fair comparisons.
Obviously, using real data can significantly boost the recognition accuracy of the evaluated models.
MGP-STR$^{\dagger}_{LFS}$ can also obtain consistent improvement ($0.43 \%$) over MGP-STR$^{\dagger}_{CFS}$, and achieve the best results on $7$ out of $8$ benchmarks.
Thus, MGP-STR$^{\dagger}_{LFS}$ establishes new SOTA results.
These comparisons clearly demonstrate the effectiveness of the proposed MGP and LFS strategies.

\begin{table}[t]\centering
\setlength{\tabcolsep}{2.5pt}
\ra{1.3}
\caption{The comparisons with other STR methods on $3$ challenging datasets. The models with $\dagger$ are trained with $200$ million iterations. For training data, ``S'' represents synthetic training datasets (MJSynth and SynthText), and ``R'' represents real training data provided by~\cite{PARSeq}.}
\label{tab:art}
\begin{tabular}{|l|c|c|c|c|c|c|c|}
\hline
\textbf{\multirow{2}{*}{Methods} } & \textbf{\multirow{2}{*}{Year}}  &\textbf{Train} &\textbf{ArT} &\textbf{COCO}  &\textbf{Uber}   &\textbf{AVG} \\
\cline{4-7}
&  &\textbf{data} &\textbf{35149} &\textbf{9825}  &\textbf{80418}   &\textbf{125392} \\
\hline
ABINet++~\cite{ABInetv2}  & TPAMI (2023) &S &65.4 &57.1 &34.9	&45.2 \\
PARSeq~\cite{PARSeq} & ECCV (2022) &S &69.1 &60.2	&39.9	&49.7 \\
MATRN~\cite{matrn} & ECCV (2022) &S &68.9	&64.0	&40.1	&50.0 \\
MGP-STR$^{\dagger}_{CFS}$ & Ours &S &70.38 &66.20 &42.18	&51.96 \\
MGP-STR$^{\dagger}_{LFS}$ & Ours &S &70.67 &66.94 &42.59	&52.37 \\
\hline
ABINet++~\cite{ABInetv2}  & TPAMI (2023) &R &81.2 &76.4 &71.5	&74.6 \\
PARSeq~\cite{PARSeq} & ECCV (2022) &R &83.0 &77.0	&82.4	&82.1 \\
MGP-STR$^{\dagger}_{CFS}$ & Ours &R &83.48 &78.43 &82.51	&82.46 \\
MGP-STR$^{\dagger}_{LFS}$ & Ours &R &84.10 &79.81 &83.32	&83.26 \\
\hline
\end{tabular}
\vspace{-4mm}
\end{table}

\subsubsection{Results on More Challenging Datasets} 

To further verify the superiority of MGP-STR in dealing with 
challenging scenarios, we also evaluate it on $3$ large-scale datasets with real-world complexity (\ie, ArT~\cite{ArT}, COCO-Text~\cite{COCO} and Uber-Text~\cite{Uber}).
We construct $2$ types of experiments:
(1) The evaluated models are trained on synthetic training datasets (\ie, MJSynth and SynthText), and tested on the three challenging datasets (\ie, ArT, COCO-Text, and Uber-Text);
(2) The models are directly trained on the real training data as mentioned in Sec.~\ref{Sec:data}, and also evaluated on the three datasets.
The results with these $2$ different experimental settings are shown in Tab.~\ref{tab:art}.

For the experiments training on synthetic data,
the word level recognition accuracies of all methods on these challenging datasets are obviously lower than those of the standard benchmarks in Tab.~\ref{tab:sota}. This indicates that for STR methods there is still a significant gap between training on synthetic data and real data.
Typically, MGP-STR$^{\dagger}_{LFS}$ can also surpass MGP-STR$^{\dagger}_{CFS}$ and achieve the best performance with synthetic data training.
Note that the model weights of MGP-STR in Tab.~\ref{tab:art} are the same as those in Tab.~\ref{tab:sota}.
These experiments show that MGP-STR has excellent robustness and generalization ability.

Compared with training on synthetic data, when training on real datasets, the overall accuracy of all methods has been greatly improved (about $30 \%$ absolute gain in performance) on challenging datasets, while only about $2 \%$ improvements on standard benchmarks in Tab.~\ref{tab:sota}.
This proves that using real data can mitigate the impact of different data distributions, and attain much higher accuracy for real challenging images.
Moreover, MGP-STR$^{\dagger}_{LFS}$ consistently achieves the best results, compared with the recent STR methods, confirming that MGP-STR can adapt well to scene text images with real-world challenges.

\begin{table}[t]\centering
\setlength{\tabcolsep}{4pt}
\ra{1.3}
\caption{The comparisons with other STR methods on three handwritten text benchmarks. LFS$^\#$ indicates that the LFS module is trained with MJSynth and SynthText, besides the handwritten text datasets. }
\label{tab:hand}
\begin{tabular}{|l|c|c|c|c|c|}
\hline
\textbf{\multirow{2}{*}{Methods} }& \textbf{\multirow{2}{*}{Year}} &\textbf{IAM} &\textbf{CVL}  &\textbf{RIMES}  \\
\cline{3-6}
&  &\textbf{13752} &\textbf{12012}  &\textbf{7776}  \\
\hline
SeqCLR~\cite{seqclr}  & CVPR (2021) &79.9 &77.8 &92.4	\\
TextAdaIN~\cite{TextAdaIN} & ECCV (2022) &87.3 &78.2 &94.4 \\
PerSec-ViT~\cite{liu2022perceiving} & AAAI (2022) &83.7	&82.9	&-	 \\
DiG-ViT-Base~\cite{DiG} & MM (2022) &87.0 &91.3	&-	\\
\hline
MGP-STR$_{Char}$ & Ours &90.55 &89.02  &93.80	\\
MGP-STR$_{BPE}$&  Ours &86.95 &89.98 &92.40	\\
MGP-STR$_{WP}$ & Ours &84.16 &89.24 &91.78 \\
\cline{1-5}
MGP-STR$_{CFS}$   & Ours  &91.89 &90.84  &94.87 \\
MGP-STR$_{LFS}$ & Ours &90.21 &91.24 &94.30 \\
MGP-STR$_{LFS^\#}$& Ours &92.93 &91.96 &95.20 \\
\hline
\end{tabular}
\vspace{-3mm}
\end{table}

\begin{table}[t]\centering
\setlength{\tabcolsep}{2.5pt}
\ra{1.2}
\caption{Comparisons on inference time and model size.}
\label{tab:speed}
\begin{tabular}{|l|l|l|}
\hline
\textbf{Methods} & \textbf{Time (ms)} & \textbf{Params ($\times 10^6$)}   \\
\hline
ABINet-S-iter1/iter2/iter3 & 13.7/18.6/24.3 & 32.8 \\
ABINet-L-iter1/iter2/iter3 & 16.1/21.4/26.8 & 36.7 \\
MATRN-iter1/iter2/iter3 & 17.9/26.5/35.1 & 44.2 \\
\hline
MGP-STR$_{Vision}$-tiny/small/base  & 10.6/10.8/10.9  & 5.4/21.4/85.5 \\
MGP-STR$_{CFS}$-tiny/small/base  & 12.0/12.2/12.3  & 21.0/52.6/148.0 \\
MGP-STR$_{LFS}$-base  & 15.3  & 170.6 \\
\hline
\end{tabular}
\end{table}

\subsubsection{Results on Handwritten Text Datasets} 

In order to validate the adaptability of MGP-STR, we carry out experiments and comparisons on $3$ widely-used handwritten text benchmarks, and the word-level accuracies are reported in Tab.~\ref{tab:hand}.
SeqCLR~\cite{seqclr} and TextAdaIN~\cite{TextAdaIN} are pre-trained and fine-tuned with only handwritten text images.
Since the number of images in the handwritten dataset is relatively small, PerSec~\cite{liu2022perceiving} and DiG~\cite{DiG} utilize large-scale unlabeled text images for model pre-training, resulting better performance.

For better results of handwritten text recognition, we directly fine-tune the model of MGP-STR$^{\dagger}_{CFS}$ trained on synthetic datasets in Tab.~\ref{tab:sota} with $300,000$ iterations.
LFS modules are further trained on handwritten datasets.
In general, MGP-STR$_{CFS}$ can effectively fuse the multi-granularity predictions, and outperform any single classification heads.
MGP-STR$_{LFS}$ can surpass MGP-STR$_{CFS}$ on CVL, but obtain worse results on IAM and RIMES.
We suspect that since there are only a limited number of words in the handwritten text datasets, the learning of textual information is insufficient.
Thus, besides handwritten data, we add synthetic data (\ie, MJSynth and SynthText) into the LFS learning.
The resultant model, MGP-STR$_{LFS^\#}$, can achieve considerable improvements over MGP-STR$_{LFS}$ ($2.7 \%$ on IAM, $0.7 \%$ on CVL and $0.9 \%$ on RIMES).
Finally, MGP-STR$_{LFS^\#}$ achieves SOTA performances on benchmarks for handwritten text recognition, verifying the effectiveness and generalization ability of our method.

\subsubsection{Comparisons of Model Size and Inference Time} 

The model sizes and latencies of the proposed MGP-STR with different settings as well as those of ABINet, MATRN are depicted in Tab.~\ref{tab:speed}~\footnote{All the evaluations are conducted on a NVIDIA V100 GPU.}. Since MGP-STR is equipped with a regular Vision Transformer (ViT) and involves no iterative refinement, the inference speed of MGP-STR is very fast. With the ViT-Base backbone, MGP-STR$_{CFS}$ and MGP-STR$_{LFS}$ only take $12.3$ ms and $15.3$ ms, respectively. Compared with ABINet and MATRN, MGP-STR runs much faster ($15.3$ms vs. $26.8$ms), and obtains higher performance. The model size of MGP-STR is relatively large. However, a large portion of the model parameter is from the BPE and WordPiece branches. Meanwhile, the actual GPU memory usage of MGP-STR$_{LFS}$-Base is small (about $2$ GB), which allows for the deployment on commonly-used devices. For scenarios that are sensitive to model size or with limited GPU memory, MGP-STR$_{Vision}$ is an excellent choice; for scenarios where resources (compute and memory) are sufficient, MGP-STR$_{LFS}$ is recommended.

\subsection{Qualitative Results}  
\label{Sec:Qualitative}

\begin{figure}[t]\centering
 \includegraphics[width=0.48\textwidth]{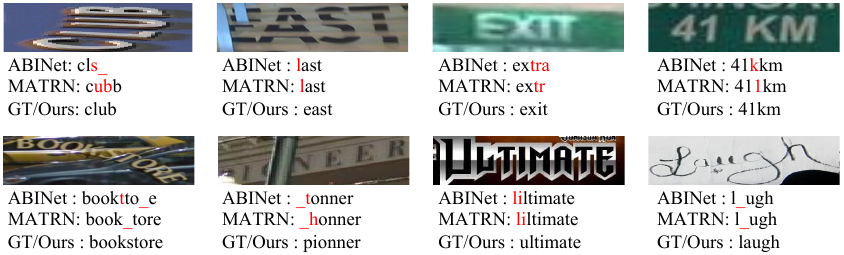}
 \caption{Qualitative comparisons with ABINet and MATRN on several typical examples.}
 \label{fig:vis_matrn}
\vspace{-3mm}
\end{figure}

\begin{figure}[t]\centering
 \includegraphics[width=0.48\textwidth]{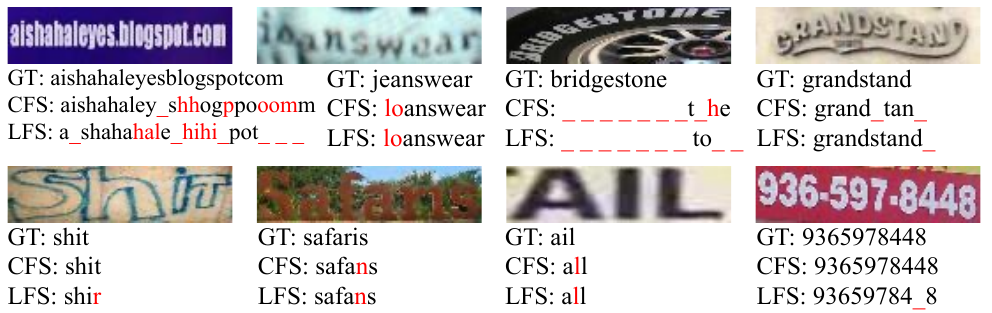}
 \caption{Failure examples of MGP-STR. In the first row, all the multi-granularity predictions are incorrect. In the second row, there are correct predictions in multi-granularity results, but the results of MGP-STR$_{LFS}$ are wrong.}
 \label{fig:vis_bad}
\end{figure}

\begin{figure}[t]\centering
 \includegraphics[width=0.48\textwidth]{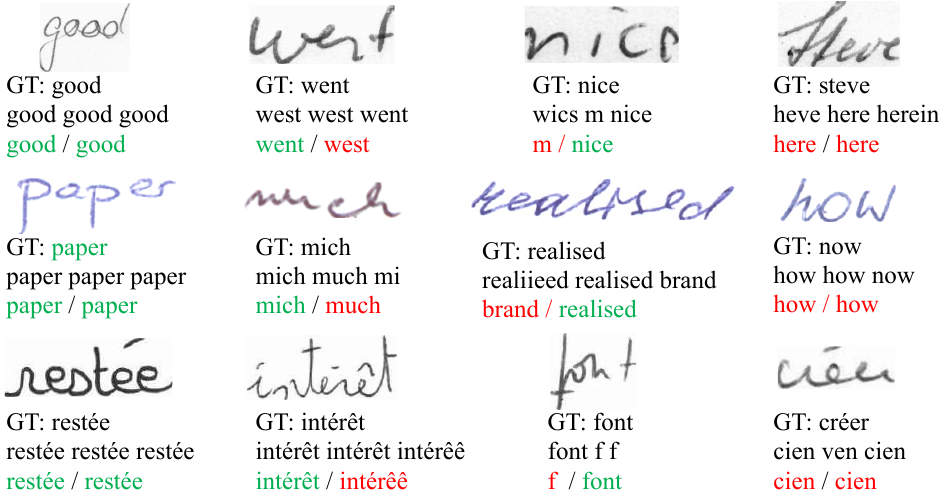}
 \caption{Visualization results of MGP-STR on $3$ handwritten text datasets. The images in the first/second/third
 row are from IAM / CVL / RIMES. The results of each image are ground-truth, multi-granularity predictions (Char BPE WP), and fusion results (CFS / LFS), respectively. (Best viewed in color.)}
 \label{fig:vis_hand}
\vspace{-4mm}
\end{figure}

We conduct qualitative comparisons with two representative STR methods (\ie ABINet~\cite{ABInet} and MATRN~\cite{matrn}) on typical images from standard benchmarks.
Fig.~\ref{fig:vis_matrn} shows the examplar images on which both ABINet and MATRN fail but MGP-STR$_{LFS}$ succeeds.
Clearly, MGP-STR$_{LFS}$ can produce correct results on vertical text, curved text, occluded text, text with irregular font, and handwritten text.
Moreover, MGP-STR$_{LFS}$ is more robust to disturbances from the background, while ABINet and MATRN tend to predict redundant characters, such as the extra ``k'' and ``1'' in the ``41km'' image.

We also illustrate typical failure cases of MGP-STR in Fig.~\ref{fig:vis_bad}. As can be seen, in the first row, MGP-STR$_{CFS}$ is not good at dealing with extremely long and heavily curved text (with complex background).
This is probably because they are rare in the training set so that the A$^3$ modules are not sufficiently learned.
Moreover, extremely vague text and words with artistic styles may pose challenges.
For the first three images in the second row, MGP-STR with LFS gave wrong predictions, due to the ambiguous visual clues in the images.
For the last image with long digits, MGP-STR$_{CFS}$ can provide correct results, while MGP-STR$_{LFS}$ mode wrong fusion decision, probably due to the weak linguistic prior in combinations of pure digits.

We also show the qualitative results of handwritten text images in Fig.~\ref{fig:vis_hand}.
Compared with scene text, handwritten text is more difficult, due to flexible writing styles, cursive writing, and special characters.
Specifically, images with regular text in the first column can be successfully recognized.
In the second column, the results of MGP-STR$_{CFS}$ are correct, while MGP-STR$_{LFS}$ are wrong.
In the third column, MGP-STR$_{LFS}$ can produce right results while MGP-STR$_{CFS}$ failed, revealing that the latter is not good at predicting correct text lengths for challenging cases.
For the last column, since the images are with ambiguous text instances, MGP-STR gave incorrect results.

\begin{figure*}[t]\centering
 \includegraphics[width=0.95\textwidth]{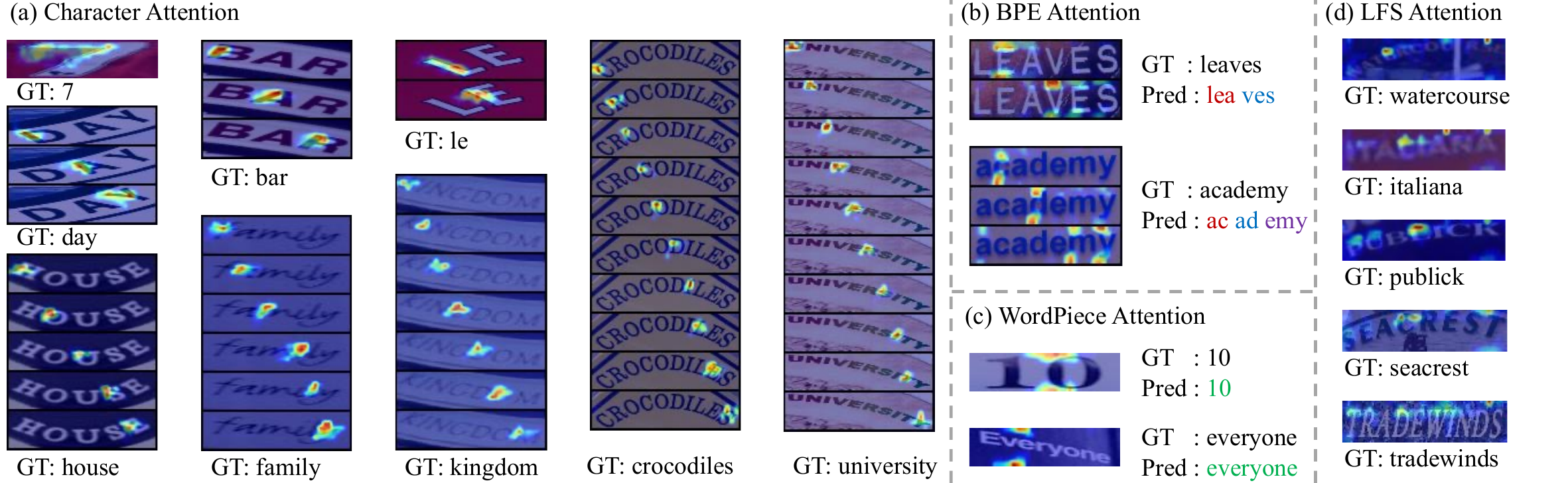}
 \caption{Illustration of spatial attention maps on (a) Character A$^3$  module, (b) BPE A$^3$  module, (c) WordPiece A$^3$  module, and (d) Image A$^3$ module of LFS respectively. (Best viewed in color.)}
 \label{fig:viscase}
\vspace{-3mm}
\end{figure*}

\subsection{Visualization of Attention Maps of A$^3$ Modules} 

Exemplar attention maps $\mathbf{m}_i $ of the Character, BPE, WordPiece, and Image A$^3$ modules are shown in Fig.~\ref{fig:viscase}.
The Character A$^3$ module shows extremely precise addressing ability on a variety of text images.
Specifically, for the ``7'' image with one character, the attention mask seems like the ``7'' shape.
For the ``day'' and ``bar'' images with three characters, the attention masks of the middle character ``a'' are completely different.
In addition, the character attention maps can perform well on curved, hand-written, and even long-curved images.
These qualitative analyses verify the capability of the Character A$^3$ module for adaptive addressing and aggregation.

The BPE A$^3$ module tends to generate short segments, as depicted in Fig.\ref{fig:mgp} and in Tab.~\ref{tab:fa}. The attention masks of BPE are spilt into $2$ or $3$ areas as shown in ``leaves'' and ``academy'' images.
Due to the fact that performing subword splitting and character addressing simultaneously may be difficult, the attention masks of the BPE A$^3$ module are not as such precise as those of the Character A$^3$ module.
Moreover, the WordPiece A$^3$ module often produces a whole word, and the attention maps are expected to cover the whole feature map. However, in reality, they are usually sparse, because the softmax function is used to these attention maps.
These results are consistent with those of Tab.~\ref{tab:CBW}, where the accuracies of ``BPE'' and ``WP'' are relatively lower than ``Char'', due to the difficulty of precise subword prediction. 

The attention maps of the Image A$^3$ module in LFS are also sparse. However, the activations appear at certain character areas that might play an important role in computing text-image similarities. Therefore, we think the Image A$^3$ module can be regarded as an attention pooling module for the whole image representation.

\section{Discussion}
\label{sec:discussion}

\subsection{Reflections}

In this section, we will discuss possible reasons for the effectiveness of the proposed Multi-Granularity Prediction (MGP) strategy and the Learnable Fusion strategy (LFS).

As demonstrated in Sec.~\ref{Sec:sota}, the proposed MGP-STR, even without an explicit language model, can outperform previous SOTA STR methods as well as a ViT-based baseline. We conjecture that: (1) One reason might be that the three output branches (Character, BPE and WordPiece), realized via multi-task learning, can capture the features from different aspects and are highly complementary. (2) Another reason, which we think is more essential, is that \textit{\textbf{language, as a kind of compositional objects, possesses the nature of multi-granularity}} and the Multi-Granularity Prediction strategy conforms well to this nature, thus leading to higher text recognition accuracy. (3) Last but not least, the subword branches (BPE and WordPiece), working at a more macroscopic level, are much robuster to difficult cases where blur, deformation and occlusion occur, compared with the conventional character-level representation.

Regarding the fusion part, the proposed LFS actually brings in a ``look-back’’ mechanism, \ie, taking a \textit{backward} pathway to verify the matching degrees between the predicted text (words) and the input image after the \textit{forward} pass, which makes the proposed algorithm go beyond a pure discriminative model, to certain extent partially similar to the ``analysis-by-synthesis’’ paradigm~\cite{yuille2006vision,bever2010analysis}.

\subsection{Limitations and Future Works}

Though introduced new strategies and achieved competitive performance, MGP-STR still has several drawbacks which need to be addressed in the future: (1) Limited by the maximum sequence length $T$ (which is set to $27$ currently), MGP-STR can only correctly read text shorter than this threshold. We will extend MGP-STR to handle longer words. (2) Even running quite fast, the model sizes of the MGP-STR series are relatively large, hindering the applicability range of MGP-STR (especially for mobile devices and embedded systems). Reducing the model parameters while keeping the recognition accuracy will be another direction for improvement. (3) Due to the adopted tokenizers and training data, MGP-STR cannot recognize text other than English and Arabic numerals. In the future, we will construct a unified text recognizer, which is able to support more types of languages (such as Chinese, Spanish, Arabic and German).

\section{Conclusion}

In this paper, we first presented a ViT-based pure vision STR model, which shows its superiority in recognition accuracy. To improve this baseline model, we then proposed a Multi-Granularity Prediction strategy to implicitly make use of linguistic knowledge. In particular, subword representations (BPE and WordPiece) are utilized to induce linguistic information in an implicit manner through multi-task learning and optimization. The proposed algorithm (named MGP-STR) with Confidence-based Fusion Strategy (CFS) is a conceptually simple yet powerful STR model without an independent language model. In addition, we propose a Learnable Fusion Strategy (LFS) to directly measure the similarities between words and images to unlock the potential of the multi-granularity predictions, further boosting the performance. MGP-STR with LFS pushes the recognition accuracy on standard datasets to a new height.

Extensive experiments are conducted to demonstrate the superiority of MGP-STR. Concretely, MGP-STR achieves state-of-the-art performances on six widely-used benchmarks (IC13~\cite{IC13}, SVT~\cite{SVT}, IIIT~\cite{IIIT}, IC15~\cite{IC15}, SVTP~\cite{SVTP} and CUTE~\cite{CUTE}) and standard handwritten text benchmarks (IAM~\cite{IAM}, CVL~\cite{CVL} and RIMES~\cite{RIMES}). Besides, when using synthetic data or real data for training, MGP-STR can also obtain state-of-the-art accuracy on recent challenging scene text datasets (ArT~\cite{ArT}, COCO-Text~\cite{COCO} and Uber-Text~\cite{Uber}). In the future, we will extend the idea of multi-granularity prediction with learnable fusion to broader domains and more tasks.



\ifCLASSOPTIONcompsoc
  \section*{Acknowledgments}
\else
  \section*{Acknowledgment}
\fi

The authors would like to thank the reviewers for their hard work and valuable suggestions.

\ifCLASSOPTIONcaptionsoff
  \newpage
\fi

\bibliographystyle{IEEEtran}
\bibliography{MGP-STR}

\begin{IEEEbiography}[{\includegraphics[width=1in,height=1.25in,clip,keepaspectratio]{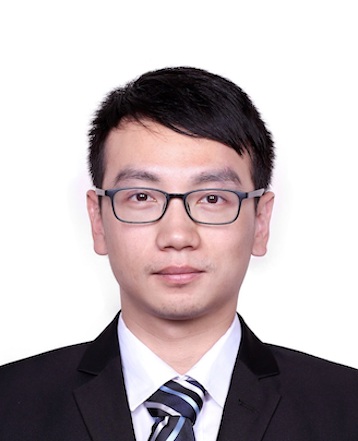}}]{Cheng Da}
received the B.S. degree in computer science from Sichuan University, Chengdu, China, in 2014, and the Ph.D. degree in National Laboratory of Pattern Recognition, Institute of Automation, Chinese Academy of Sciences, Beijing, China, in 2019. He is currently an algorithm researcher at Alibaba DAMO Academy, Beijing, China. His current research interests include scene text recognition and computer vision.
\end{IEEEbiography}

\begin{IEEEbiography}
[{\includegraphics[width=1in,height=1.25in,clip,keepaspectratio]{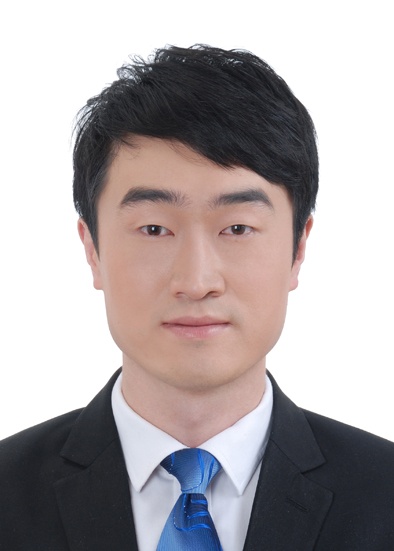}}]{Peng Wang}
received the B.S. degree from the National Key Laboratory of Science and Technology on Communications at Beijing University of Posts and Telecommunications, Beijing China, in 2016. He is currently an algorithm researcher at Alibaba DAMO Academy, Beijing, China. His current research interests include computer vision and scene text recognition.
\end{IEEEbiography}


\begin{IEEEbiography}[{\includegraphics[width=1in,height=1.25in,clip,keepaspectratio]{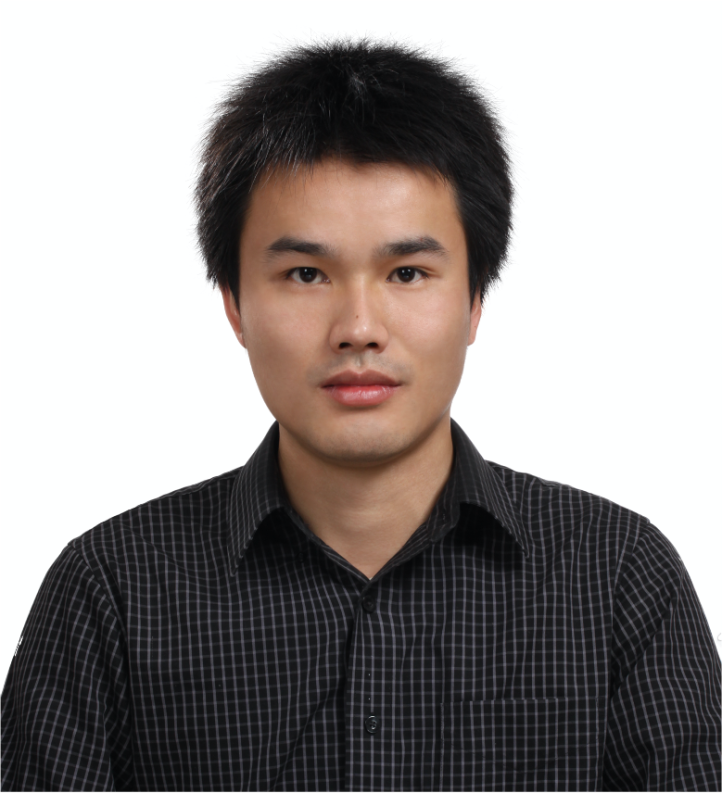}}]{Cong Yao} is currently with Alibaba DAMO Academy,
Beijing, China. He received the B.S. and Ph.D. degrees in electronics and information engineering from the Huazhong University of Science and Technology (HUST), Wuhan, China, in 2008 and 2014, respectively. He was a research intern at Microsoft Research Asia (MSRA), Beijing, China, from 2011 to 2012. He was a Visiting Research Scholar with Temple University,
Philadelphia, PA, USA, in 2013. His research has focused on computer vision and deep learning, in particular, the areas of scene text reading and document understanding.
\end{IEEEbiography}




\end{document}